%% file: main.tex
\documentclass{article}
\usepackage{mathpazo}
\usepackage[margin=1in]{geometry}
\usepackage{hyperref}
\usepackage{amsmath,amssymb,amsthm}
\usepackage{mathtools}
\usepackage{bm}
\usepackage{caption}
\usepackage[capitalize,noabbrev]{cleveref} %
\crefformat{section}{\S#2#1#3} %
\crefformat{subsection}{\S#2#1#3}
\crefformat{subsubsection}{\S#2#1#3}

\usepackage{comment}

\usepackage{textcomp}
\usepackage{tikz}
\usepackage{graphicx}
\usepackage{forest}
\usepackage{tikz}
\usepackage{amssymb}%
\usepackage{pifont}%
\newcommand{\dtarg}{D}
\newcommand{\st}[1]{\mathbf{s}_{#1}}

\usepackage{tikz-cd}
\usepackage{pgf}
\usepackage{pgfplots}
\usepackage{pgfplotstable}
\pgfplotsset{compat=newest}
\usepgfplotslibrary{groupplots,colorbrewer,fillbetween,colormaps}
\usetikzlibrary{plotmarks} %
\usetikzlibrary{backgrounds}
\usetikzlibrary{arrows.meta}

\makeatletter
\pgfmathdeclarefunction{strcmp}{2}{%
  \begingroup
  \edef\pgfmath@tempa{#1}%
  \edef\pgfmath@tempb{#2}%
  \pgfmathparse{%
    \ifnum\pdfstrcmp(\pgfmath@tempa,\pgfmath@tempb)==0 0\else 1\fi%
  }%
  \endgroup
}
\makeatother

\DeclareMathOperator*{\argmin}{arg\,min}

\newtheorem{definition}{Definition}
\newtheorem{remark}{Remark}

\usepackage{nicefrac}
\usetikzlibrary{decorations.pathreplacing} %
\usetikzlibrary{positioning,arrows.meta,decorations.pathreplacing}
\usetikzlibrary{patterns}
\usetikzlibrary {patterns,patterns.meta}
\usetikzlibrary{shapes.geometric}
\usetikzlibrary{fit,calc}

\colorlet{mypink}{red!40}
\colorlet{myblue}{cyan!60}

\definecolor{commentcolor}{RGB}{77, 77, 77}
\definecolor{forcolor}{RGB}{203, 49, 56}
\definecolor{fncolor}{RGB}{93, 48, 175}

\usepackage{xspace}
\newcommand{\replay}{\textsc{Replay}\xspace}

\tikzset{
  simple cylinder/.style={
    shape=cylinder,
    draw,
    shape aspect=0.2,
    cylinder uses custom fill,
    shape border rotate=90,
    transform shape,
  }
}

\DeclareRobustCommand\tikzinlinecylinder{\raisebox{-0.05em}{\tikz
\node[simple cylinder, fill=gray!20,inner sep=3.2pt,outer sep=3.2pt] {};}}

\newcommand{\pfrac}[2]{\frac{\partial #1}{\partial #2}}

\usepackage[linesnumbered,ruled,vlined]{algorithm2e}
\let\oldtcp\tcp

\let\oldnl\nl
\newcommand{\nonl}{\renewcommand{\nl}{\let\nl\oldnl}}

\renewcommand{\tcp}[1]{\oldtcp{\textcolor{commentcolor}{#1}}}
\renewcommand{\epsilon}{\varepsilon}

\usepackage{nicefrac}
\usepackage{color-edits}
\usepackage{enumitem}
\usepackage{amsfonts} %
\DeclareMathSymbol{\shortminuss}{\mathbin}{AMSa}{"39}
\newcommand{\shortminus}[1]{-} %

\addauthor[Andrew]{ai}{blue}

\usepackage[backend=biber,style=alphabetic,natbib=true,maxbibnames=99,minalphanames=3,maxcitenames=2]{biblatex}
\usepackage[T1]{fontenc}
\usepackage{booktabs}
\usepackage[colorinlistoftodos,textsize=tiny,textwidth=2.1cm,shadow,loadshadowlibrary]{todonotes}

\usepackage{appendix}
\usepackage{subcaption}

\definecolor{mygreen}{rgb}{0.00784313725490196, 0.6196078431372549,
0.45098039215686275}
\definecolor{myred}{rgb}{0.8352941176470589, 0.3686274509803922, 0.0}
\definecolor{mydarkblue}{rgb}{0,0.08,0.85}
\definecolor{mylightblue}{rgb}{0.06,0.56,1.0}
\definecolor{mylightorange}{rgb}{1.0,0.62,0.12}
\definecolor{mylightred}{rgb}{0.99,0.00,0.04}
\definecolor{pgfcolor0}{HTML}{66c1a5}
\definecolor{pgfcolor1}{HTML}{fc8c61}
\definecolor{pgfcolor2}{HTML}{8ca0cb}
\definecolor{pgfcolor3}{HTML}{e68ac3}
\hypersetup{colorlinks=true,
  linkcolor=mydarkblue,
  citecolor=mydarkblue,
  filecolor=mydarkblue,
  urlcolor=mydarkblue,
  bookmarksopen=true,
linktoc=page}

\usepackage[T1]{fontenc}
\title{Optimizing ML Training with Metagradient Descent}

\author{
    Logan Engstrom${}^{*\,1}$,
    Andrew Ilyas${}^{*\,2}$\stepcounter{footnote}\footnote{Work done at MIT EECS. Correspondence to \texttt{\{engstrom,ailyas,benchen\}@mit.edu}.}\ ,
    Benjamin Chen${}^{*\,1}$, \\
    Axel Feldmann${}^{1}$\ ,
    William Moses${}^{3}$,
    Aleksander M\k{a}dry${}^{1}$ \\[.5em]
    {\normalsize $^*$Equal contribution \qquad $^1$MIT,\ $^2$Stanford,\ $^3$UIUC}
}
\date{}

\begin{document}
\maketitle
\vspace{-1em}
\begin{abstract}
    A major challenge in training large-scale machine learning models 
    is {\em configuring} the training process to maximize model performance,
    i.e.,
    finding the best training setup from a vast design space. 
    In this work, we unlock a gradient-based approach to this problem.
    We first introduce an algorithm for efficiently calculating 
    \textit{metagradients}---gradients through model training---at scale.
    We then introduce a ``smooth model training'' framework
    that enables effective optimization using metagradients.
    With metagradient descent (MGD), we greatly improve on existing
    dataset selection methods, outperform accuracy-degrading
    data poisoning attacks by an order of magnitude, 
    and automatically find competitive learning rate schedules.
\end{abstract}

\section{Introduction}
\input{sections/intro}

\section{Scalably computing metagradients}
\label{sec:computing}
\input{sections/altalt_method.tex}

\section{Designing metasmooth training routines} %
\label{sec:localpred}
\input{sections/localpred}

\section{Applications}
\label{sec:apps}
\input{sections/applications.tex}

\section{Discussion}
\label{sec:discussion}
\input{sections/discussion.tex}

\section{Related work}
\label{sec:related}
\input{sections/ext_related.tex}

\section{Conclusion}
\label{sec:conclude}
\input{sections/conclusion.tex}

\section{Acknowledgements}
Work supported in part by the NSF grant DMS-2134108 and Open Philanthropy,
and in part by NSF Grant No. 2346519. 
This work is also supported in part by the
Alan Turing Institute, and the U.S. Department of Energy.
The authors would like to thank 
Alex Damian,
Harshay Shah,
Jesse Michel,
Joel Flynn,
Manolis Zampetakis,
Noah Moroze,
Piotr Indyk,
Sam Hopkins,
Sung Min (Sam) Park,
and
Sarah Cen
for helpful references as well as 
discussions and feedback on early versions of this work.

\clearpage
\printbibliography

\appendix

\clearpage
\onecolumn

\clearpage
\section{Calculating metagradients with \replay{}}
\input{sections/appmethods.tex}

\clearpage
\section{Smooth Model Training}
\subsection{Omitted Figures}
\begin{figure}[h]
  \centering
  \includegraphics[width=.46\textwidth]{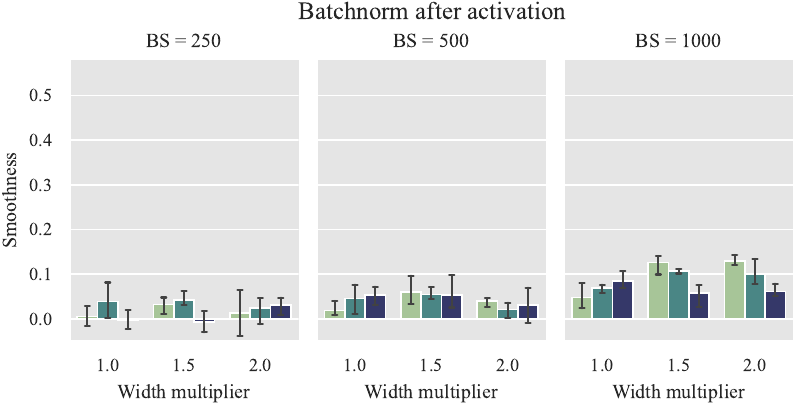}
  \includegraphics[width=.51\textwidth]{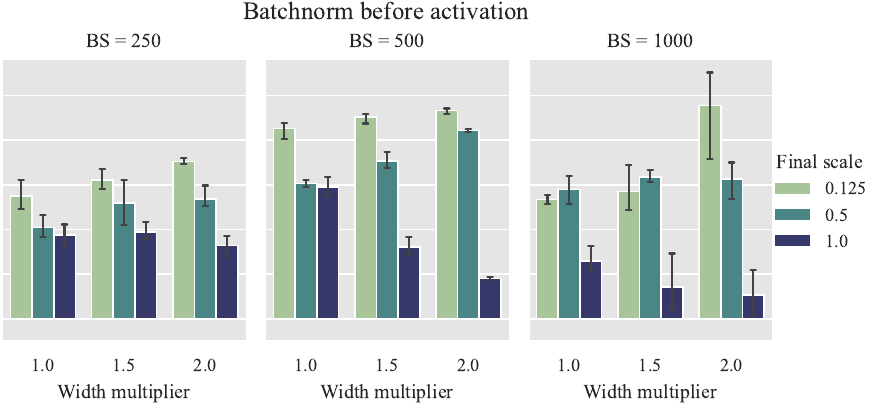}
  \caption{The factors affecting metasmoothness of training a ResNet-9 on the
  CIFAR-10 dataset. See \cref{sec:localpred} for details.}
  \label{fig:smoothness_cifar}
\end{figure}

\begin{figure}
    \centering 
\newcommand{\plotnumber}{1118}
\newcommand{\zmin}{-1}
\newcommand{\zmax}{11}
\newcommand{\zdist}{3}
\input{pgffigs/smoothness_3d/smoothness_3d.tex}

\vspace{.3cm}
\renewcommand{\plotnumber}{3349}
\renewcommand{\zmin}{-1}
\renewcommand{\zmax}{11}
\renewcommand{\zdist}{3}
\input{pgffigs/smoothness_3d/smoothness_3d.tex}

\vspace{.3cm}
\renewcommand{\plotnumber}{10600}
\renewcommand{\zmin}{-1}
\renewcommand{\zmax}{9}
\renewcommand{\zdist}{3}
\input{pgffigs/smoothness_3d/smoothness_3d.tex}

\vspace{.3cm}
\renewcommand{\plotnumber}{15578}
\renewcommand{\zmin}{-1}
\renewcommand{\zmax}{9}
\renewcommand{\zdist}{3}
\input{pgffigs/smoothness_3d/smoothness_3d.tex}
\caption{Additional loss landscape visualizations.}
\label{fig:more_landscapes}
\end{figure}

\clearpage
\section{Metagradients for DataComp}
\label{app:clip_details}
\input{sections/app/clip_details.tex}

\clearpage
\section{Selecting IFT data}
\label{app:less}
\input{sections/app/ift_details.tex}

\clearpage
\section{Accuracy-degrading data poisoning}
\label{app:poisoning}

\subsection{Background on Gradient Cancelling attack}
We briefly review the Gradient Cancelling attack \citep{lu2023exploring}
used as a baseline in our experiments. We refer the reader to the original
paper for details. Here we highlight the key ideas.

At a high level: Gradient Cancelling (GC) explicitly aims at making a specific
malicious parameter configuration reachable through retraining on the poisoned
dataset. The attack operates in two phases:

\begin{enumerate}
  \item \textbf{Parameter Generation}: The attacker generates a target malicious model
  parameter independently, often using a direct parameter corruption method like
  Gradient-based Parameter Corruption (GradPC) \citep{lu2023exploring}. 
  The end result of this phase is a target model parameter $\theta_p$
  that achieves low accuracy on the test set, but is close to the original
  parameter $\theta_0$ derived from training on the clean dataset.

  \item \textbf{Poison Data Crafting}: In the second phase, GC finds values of the poison 
  data that induce a near-zero gradient at the target parameter $\theta_p$.
  This is achieved by solving a gradient cancellation optimization
  problem: specifically, GC minimizes the total gradient of the loss function
  (with respect to the model parameters) evaluated over the combined (clean and
  poisoned) dataset, aiming to ensure that the gradient at the malicious parameter
  $\theta_p$ approaches zero. 
\end{enumerate}

\subsection{Metasmooth hyperparameters}
\begin{table}[h]
  \caption{Hyperparameters used in the ResNet-9 \citep{jordan202494}
    CIFAR-10 poisoning experiments.  The augmentations used are
  normalization, random horizontal flip, and random translate (2 pixels)}
  \centering
  \input{tables/cifar_hparams.tex}
\end{table}

\clearpage
\section{LR optimization}
\label{app:lr_opt}

\begin{table}[h]
  \caption{The grid search was run over all 528 combinations of the
  hyperparameter values below.}
  \centering
  \begin{tabular}{l c} \toprule
    \textbf{Parameter} & \textbf{Values} \\
    \midrule
    Peak learning rate & [7.0, 7.5, 8.0, 8.5, 9.0, 9.5, 10.0, 10.5,
    11.0, 11.5, 12.0] \\
    Initial LR multiplier & [0.05, 0.15, 0.25, 0.35, 0.45, 0.55] \\
    Final LR multiplier & [0.05, 0.15, 0.25, 0.35, 0.45, 0.55] \\
    LR peak time & [0.25, 0.5, 0.75] \\
    \bottomrule
  \end{tabular}
  \label{tab:lr_grid_hparams}
\end{table}

\begin{figure}[h]
  \centering
  \input{pgffigs/lr_sched_plot/sched_plot.tex}
  \caption{Graphs of our learned LR schedules.}
  \label{fig:lr_sched_plot}
\end{figure}

\end{document}

%% file: sections/intro.tex
{\em How should I clean my data? What architecture should I use?}
Training large-scale (i.e., deep) machine learning models entails
making many design decisions.
When making such decisions,
typical practice is to
exhaustively search over a small set of standard options.
For example, we might try a few well-known
data cleaning heuristics,
construct a grid over a hyperparameters,
and choose the options that yield the best models.
However, given that this process explores only a small part of the overall
design space (e.g., one can construct $2^n$ possible training datasets from
a pool of $n$ candidate datapoints),
it is unlikely that this approach really yields the
\textit{optimal} training configuration.

How can we find optimal (or at least, better) training configurations?
To do so,
we take the {optimization} perspective on designing model
training.
From this well-studied
perspective,
deciding on a training configuration---or as we will call it,
a set of {\em metaparameters}---is just a high-dimensional optimization
problem.
The input space of this problem comprises all possible metaparameter choices,
including which datapoints to train on, what model architecture to use,
and how to initialize model weights.
The objective function
takes in a set of metaparameters,
trains a machine learning model according to those metaparameters,
and then returns a target metric evaluated on that model
(e.g., test accuracy).
From this perspective, any procedure for selecting
metaparameters---including the typical practice of
grid-searching over standard options---is just an optimization algorithm,
whose goal is to
maximize the objective function with respect to the (high-dimensional) input.

Given that selecting metaparameters is ``just'' a
high-dimensional optimization problem,
a natural tool to consider is the {\em gradient}.
After all, in many contexts,
gradients offer a more effective approach to maximizing
high-dimensional functions than grid search.
Indeed, for a sufficiently
``well-behaved'' function $\smash{f(x)}$ with gradient $\smash{\nabla f(x)}$,
we can optimize $f$ by
iteratively updating $x$ in the direction of $\nabla f(x)$.
This insight suggests a generic recipe for selecting metaparameters:
first, make the objective differentiable with respect to
the metaparameters; second, update via gradient steps.

Now, the idea of using gradients to search for metaparameters is not new.
Indeed, there is a substantial line of work that aims to optimize metaparameters
(e.g., architectures, regularizers, or data augmentation schemes)
with gradient-based methods
\citep{maclaurin2015gradient,liu2018darts,lorraine2020optimizing}.
However, such methods have not managed to scale beyond relatively
small settings. This state of affairs prompts our main question:
\begin{center}
  \vspace*{-.2em}
  {\em Can we \underline{\smash{scalably}} configure model training
  using gradient-based methods?}
  \vspace*{-.2em}
\end{center}

\begin{figure}
  \centering
  \input{pgffigs/headline/headline}
  \caption{Our proto-algorithm, metagradient descent (MGD),
    uses gradients to achieve state-of-the-art performance across
    a variety of applications, including data selection and
  data poisoning.}
  \label{fig:headline}
\end{figure}

\subsection{Contributions}
In this work,
we answer this question in the affirmative,
adding ``gradient descent on metaparameters''
to the large-scale machine learning toolkit.
Along the way, we will face---and address---two main challenges.

First, existing methods for computing metagradients do not scale.
In response, we devise an algorithm, $\replay{}$,
that can take
metagradients in large-scale settings.
By combining reverse-mode autodifferentiation (AD)
with an efficient data structure,
\replay{} can calculate exact metagradients for models
with billions of parameters and thousands of training steps.

Second,
we find that metagradients of standard training routines
are not necessarily helpful for optimization,
which we connect to {\em non-smoothness} of the metaparameter
optimization landscape.
Borrowing tools from convex optimization,
we devise a framework for designing
``metasmooth'' training routines that {\em do} admit helpful
metagradients.

Addressing the challenges above
unlocks a simple recipe
for solving a broad range of machine learning tasks:
(a) frame the task as a continuous optimization problem over metaparameters;
(b) design a metasmooth training routine;
(c) perform metagradient descent (MGD).
Applying this recipe:
\begin{itemize}
  \item In the DataComp-\texttt{small} competition \citep{gadre2024datacomp},
  we achieve state-of-the-art pre-training data selection for CLIP
    (2x larger performance improvement than the previous
    DataComp-\texttt{small} leader \citep{ecodatum2024ecodatum});
  \item In the context of data selection for instruction tuning (as introduced by \citet{xia2024less}),
    we substantially improve on data selection for Gemma-2B
    (outperforming existing selection methods
    as well as full-data training);
    \item In the {\em accuracy-degrading} data poisoning setting
    (defined by \citet{huber1964robust} and pioneered by
    \citet{lu2022indiscriminate} for deep neural networks),
      we improve attacks on DNNs by an order of magnitude,
      dropping CIFAR-10 accuracy from $92\%\!\to\! 78\%$
      (the best previous attack \citep{lu2023exploring} 
      only reduces accuracy to $91\%$);
  \item For the task of hyperparameter optimization, we 
    efficiently find a competitive CIFAR-10 learning rate schedule 
    (matching the performance of a schedule found by grid search).
\end{itemize}

\begin{figure}[h]
  \centering
  \begin{tikzpicture}[>=stealth,font=\Large]
    \node (z) at (0,0) {$z$};
    \node (theta) at (3,0) {$\theta = \mathcal{A}(z)$};
    \node (phi) at (6,0) {$\phi(\theta)$};

    \draw[->] (z) -- (theta);
    \draw[->] (theta) -- (phi);

    \draw[blue,dashed,->] (phi) to[bend right=40] node[above]
    {$\nabla_z \phi(\mathcal{A}(z))$} (z);

    \node[above] at (0,-0.9) {\normalsize Training setup};
    \node[above] at (3,-0.9) {\normalsize Trained model};
    \node[above] at (6,-0.9) {\normalsize Observed behavior};
  \end{tikzpicture}
  \caption{An illustration of the metagradient.
    We embed a given aspect of the training setup (e.g., the training
    dataset, or optimizer hyperparameters)
    into a continuous {\em metaparameter} vector $z \in \mathbb{R}^d$.
    This metaparameter defines a model $\mathcal{A}(z)$
    by way of the learning algorithm $\mathcal{A}$,
    which in turn defines an output $\phi(z)$.
    The {\em metagradient}
    $\nabla_z \phi(\mathcal{A}(z)) \in \mathbb{R}^d$
    is the gradient of this model output with respect to
  the metaparameter.}
  \label{fig:meta_gradient}
\end{figure}

%% file: pgffigs/headline/headline.tex
\pgfplotsset{
  compat=1.17,
  cycle list/Set2,
}
\begin{tikzpicture}
  \begin{groupplot}[
      group style={
        group size=3 by 1,
        horizontal sep=1.5cm,
      },
      width=5.5cm, height=5cm,
      /pgf/bar width=1cm,
      ybar,
      xtick=data,
      enlarge x limits=0.5, %
      ymin=0,
      xlabel={Method},
      xmajorgrids=false,    %
      ymajorgrids=true,     %
      grid style={dashed},
      /pgf/bar shift=0cm,
    ]
    \nextgroupplot[
      title={CLIP Data Selection}, 
      symbolic x coords={DataComp \#1, MGD},
      ylabel={$\Delta$ over no filtering (\%)},
      xtick={DataComp \#1, MGD},
    ] 
    \addplot[ybar, fill=gray] coordinates {(DataComp \#1,5)};
    \addplot[ybar, fill=Set2-A] coordinates {(MGD,9)};
      
    \nextgroupplot[
      title={IFT Data Selection}, 
      symbolic x coords={LESS, MGD},
      xtick={LESS, MGD},
      ylabel={$\Delta$ over full dataset (\%)},
    ] 
    \addplot[ybar, fill=gray] coordinates {(LESS,0.25)};
    \addplot[ybar, fill=Set2-A] coordinates {(MGD,1.4)};
      
    \nextgroupplot[
      title={Data Poisoning}, 
      symbolic x coords={GC, MGD},
      xtick={GC, MGD},
      ylabel={Test accuracy drop (\%)},
    ] 
    \addplot[ybar, fill=gray] coordinates {(GC,0.8)};
    \addplot[ybar, fill=Set2-A] coordinates {(MGD,15.4)};
      
  \end{groupplot}
\end{tikzpicture}

%% file: sections/altalt_method.tex
In this section we present \replay{}, an algorithm for computing metagradients
of large-scale iterative ML algorithms. We first detail the setting, 
then discuss existing approaches to computing metagradients, 
and conclude by describing \replay{}.

\subsection{What is a metagradient?}
\label{subsec:meta_gradient}
Training a machine learning model is a two-step process. 
First, we decide on a {\em training setup}---we must pick,
for example, a neural network architecture, 
a training dataset, and an optimizer for training.
Second, we apply the algorithm defined by this training setup to 
train a model.

Our overall goal in this paper is to optimize model behavior as a 
function of the training setup 
(or, as we call it, the {\em metaparameters}) using
gradient-based methods. To this end, we define the following notation:
\begin{itemize}
	\item Let $\mathbf{z} \in \mathbb{R}^n$ be a vector of continuous 
    metaparameters representing the aspects of the training setup we aim to optimize. 
    For example, if we only want to adjust the learning rate and weight decay of
    SGD then $n = 2$. We handle discrete metaparameters (e.g., choice of training data)
    by finding a continuous relaxation (e.g., importance weights).
	\item Let $\mathcal{A}$ be an {\em algorithm} mapping $\mathbf{z}$ to a trained
    machine learning model; we assume all other
	aspects of the training setup outside $\mathbf{z}$ are fixed and thus part of the
    algorithm $\mathcal{A}$.  
    \item Finally, let $\phi$ be an {\em output function} mapping a model
    $\theta$ to a vector $\phi(\theta) \in \mathbb{R}$.
    For example, $\phi(\theta)$ might represent the validation loss of the model $\theta$.
    We require that $\phi$ be differentiable with respect to $\theta$, but otherwise 
    make no assumptions on $\phi$.
\end{itemize}
With this notation in place, we define the 
{\em training function} $f := \phi \circ \mathcal{A}$
mapping the training setup $\mathbf{z}$ 
{\em directly} to the output function $\phi$ evaluated on the corresponding model. 

Finally, the {\em metagradient} is the gradient of the training function with 
respect to the metaparameters, $\nabla_\mathbf{z} f(\mathbf{z})$. 
Intuitively, the metagradient defines the ``direction of steepest ascent'' in
metaparameter space.

\paragraph{Our focus: iterative algorithms.}  
To efficiently compute the metagradient, 
we restrict our focus to cases where the algorithm $\mathcal{A}$ is 
{\em iterative}, i.e., when it can be written in the form
\begin{align}
  \label{eq:iterative_algo}
  \underbrace{\mathcal{A}(z) \coloneqq \st{T}}_{\mathclap{\text{model state after $T$ steps}}}, 
  \quad \text{where} \quad \underbrace{\st{t+1} := h_t(\st{t}, \mathbf{z})}_{\mathclap{\text{optimizer step $t$}}}.
\end{align}
Here, $\st{t}$ is the optimizer state at step $t$
(with $\st{0}$ being the initial state)
and $h_t$ is the {\em update} mapping from state $t$ to state $t+1$.
The form of~\eqref{eq:iterative_algo} captures most large-scale training algorithms. 
For example, if the setup $\mathbf{z} \in \mathbb{R}^T$ is a \textit{per-step} learning rate, 
and the algorithm $\mathcal{A}$ is 
full batch gradient descent, then each update $h_t$ is
$$
  h_t(\st{t}, \mathbf{z}) \coloneqq \st{t} - z_t \nabla \ell(\st{t}),
$$
where $z_t$ is the learning rate at step $t$, 
$\ell$ is the training loss, and 
the state $\st{t}$ comprises the parameters at step~$t$.
For more complex algorithms like
Adam~\citep{kingma2015adam}, the state $\st{t}$ includes
terms like gradient moments.

\subsection{Warmup: Metagradients via autodifferentiation}
A key primitive we leverage to calculate metagradients is
{\em automatic differentiation} (AD)---a standard tool for taking 
gradients through computer-defined functions.
AD takes gradients 
by decomposing functions into elementary operations 
with known derivatives, then combining these derivatives using the chain rule.
Concretely, AD operates in two passes: a ``forward pass,'' 
which executes the function of interest and stores intermediate products 
for {each} elementary operation; and a ``backward pass,''
which calculates the gradient by propagating chains of
partial derivatives using these stored products. For the purposes of this paper, 
we will view AD as a black box that calculates the gradient of a many-to-one function 
(i.e., any $f: \mathbb{R}^d \to \mathbb{R}$)
at a given point using only a small constant factor more time than calculating the
function itself (along with the space cost of storing the 
necessary forward-pass products).

What does this have to do with metagradients? 
Well, seeing as how training itself is a computer-defined function, 
AD is a natural tool for calculating the metagradient. The main challenge, 
as we discuss in the sequel, is that AD-based approaches to calculating 
the metagradient tend to be too resource-intensive for the large-scale machine
learning algorithms we consider.
In the remainder of this section we build up background before finally
describing \replay{}, our algorithm for scalably computing (exact)
metagradients.

\paragraph{Approach \#1: Direct AD.} The direct approach to calculating
metagradients exploits the fact that nearly any learning algorithm
is itself a sequence
of differentiable computer-defined operations---meaning the training function $f$
is \emph{also differentiable}.

However, operationalizing this observation to compute metagradients
turns out to be challenging.
The reason is that AD stores intermediate
products for {\em each} operation.
The amount of data stored thus scales with the number of operations 
in the function of interest. In the case of our training function $f$,
this number encompasses {\em all} the operations used to train a machine 
learning model.
As a result, even in a toy scenario like MNIST training, computing metagradients
with na\"ive AD would require storing terabytes of data. 

\paragraph{Approach \#2: Exploiting structure with step-wise AD.} 
A more efficient method for calculating the metagradient,
{\em step-wise AD},
leverages the structure of iterative learning algorithms~\citep{werbos1990backpropagation,maclaurin2015gradient,franceschi2017forward}.
Recall from \eqref{eq:iterative_algo} that such algorithms take the form
\[
	{\mathcal{A}(\mathbf{z}) \coloneqq \st{T}}, \quad \text{where} \quad {\st{t+1} := h_t(\st{t}, \mathbf{z})}.
\]
Algebraic manipulation 
(in particular, using the chain rule, 
the law of the total derivative,
and the identity $\st{t} = h_{t-1}(\st{t-1}, \mathbf{z})$)
allows us to write the metagradient over an iterative algorithm as
\begin{equation}
	\pfrac{f(\mathbf{z})}{\mathbf{z}} =
	\pfrac{\phi(\mathcal{A}(\mathbf{z}))}{\mathbf{z}} 
	= \mathlarger{\sum}_{t=1}^{T}\ 
  \underbrace{\overbrace{\pfrac{\phi(\st{T})}{\st{t}}}^{A_t} \cdot 
  \pfrac{h_{t-1}(\st{t-1}, \mathbf{z})}{\mathbf{z}}}_{B_t} \label{eq:total_deriv}, %
\end{equation}
where we have introduced the notation $A_t$ and $B_t$ for notational convenience.
Step-wise AD computes the metagradient by calculating each term in the sum of
\eqref{eq:total_deriv} one at a time. 
For each term, the main challenge lies in computing $A_t$, since given $A_t$ 
we can straightforwardly compute $B_t$ (the entire term)
by differentiating through a single model update, i.e.,
\begin{align*}
  B_t := A_t \cdot \pfrac{h_{t-1}(\st{t-1}, \mathbf{z})}{\mathbf{z}}
  =  \pfrac{(A_t \cdot h_{t-1}(\st{t-1}, \mathbf{z}))}{\mathbf{z}},
\end{align*}
which is just a single call to our assumed ``AD oracle'' on the function 
$\mathbf{z} \mapsto A_t \cdot h_{t-1}(\st{t-1}, \mathbf{z})$.
Computing the $A_t$ terms is less straightforward as we need to relate $s_t$ and $s_T$;
to do so, we exploit the recurrence
\begin{equation}
  A_t := \pfrac{\phi(\st{T})}{\st{t}} 
  = \pfrac{\phi(\st{T})}{\st{t+1}} \cdot \pfrac{{h_t(\st{t}, \mathbf{z})}}{\st{t}}
  = \pfrac{(A_{t+1} \cdot h_t(\st{t}, \mathbf{z}))}{\st{t}},\label{eq:grad_recurrence}
\end{equation}
making $A_t$ straightforward to compute (again, a single ``AD oracle'' call) given $A_{t+1}$. 
Step-wise AD exploits this fact to successively
calculate the gradient with respect to each state, from state $T$ down to state $0$.

Bringing these ingredients together, the algorithm executes as follows. As a
preprocessing step, it trains the model and stores all intermediate states
$\st{0},\ldots,\st{T}$.
Then, the algorithm calculates and sums the terms in
\eqref{eq:total_deriv}. It first computes $A_T := \nicefrac{\partial
\phi(\st{T})}{\st{T}}$, the gradient of the output function $\phi$ with respect
to the final state. Then, the algorithm steps through $\st{T-1},\ldots,\st{0}$ in
reverse order, calculating (a) the gradient with respect to each
state $A_t$ (via \eqref{eq:grad_recurrence}) and (b) the gradient with respect to
$\mathbf{z}$ at that step $B_t$ (via \eqref{eq:total_deriv}, using the
previously calculated gradient with respect to that state). AD calculates
both quantities---each requires differentiating over only one train step.
Finally, the algorithm returns the final metagradient as the sum of the terms.

Despite improving storage overhead compared to ``direct AD'', step-wise
AD is still too space-intensive at scale. After all,
this algorithm saves {\em every} optimizer state. 

\begin{figure}
	\onecolumn
    \centering
    \input{figs/tree.tex}
    \caption{The lazy $k$-ary tree structure for
    traversing optimizer states in reverse order, with $k=2$. Recall that $n$ is
    the number of states (parameterized such that $n=T+1$). Each node represents
    the correspondingly numbered state. We give an example of the traversal
    using the \color{blue}{blue arrows }\color{black} in the figure, which
    denote the traversal path up to state $\smash{s_{\frac{3n}{4} + 1}}$. The gray
    cylinders \hspace{1pt}\tikzinlinecylinder{}\hspace{1pt} indicate the states
    that are stored when the traversal is at state $\smash{s_{\frac{3n}{4} + 1}}$; the
    other states are not stored at this point in the traversal. Traversing this
    structure requires storing $\mathcal{O}(\log(n))$ state and computing
    $\smash{\mathcal{O}(n\log(n))}$ optimizer steps---compared to $n$ for simply
    training. }
    \label{fig:efficient_data_structure}
\end{figure}

\subsection{\replay{}} 
\replay{} is our algorithm for efficiently and exactly computing metagradients.
It uses $\mathcal{O}(k\log_k(T))$ space and requires running the learning algorithm $\mathcal{A}$
a total of $1 + \log_k(T)$ times, 
with $k$ a user-chosen constant. The main idea is to make the space-intensive subroutine of
step-wise AD---a reverse-order traversal of the optimizer states at each
step---much more efficient. After all, step-wise AD stores \textit{all} the
states to reverse traverse them. \replay{} modifies step-wise AD to traverse
states in less space by exploiting a simple observation: when training is
deterministic, one can \textit{reinstantiate} an optimizer state $\st{t}$ by
``replaying'' training from a fixed point $t' < t$---at the compute cost of $t -
t'$ training steps. For example, one simple scheme saves every other state, then
``replays'' the remaining states when (reverse) traversing; this routine
stores $T/2$ states but computes an extra $T/2$ model updates compared
to storing \textit{all} the states.

\replay{} performs a reverse-order traversal the optimizer states while
balancing  the compute cost of ``replaying'' training with the storage cost of
saving states. We use a combination of deterministic training 
(fixing data ordering, data augmentation, and any other randomness in 
the training process) and an efficient data structure (similar to a segment
tree; see Figure \ref{fig:efficient_data_structure}) 
to reverse-order traverse the optimizer states with
$\mathcal{O}(k\log_k(T))$ space and an additional $T\log_k(T)$
model steps. 

Specifically, \replay{} recursively saves and replays training states. The
algorithm splits the training trajectory into $k$ segments, performs the full
training routine while saving only the start of each segment, then recurses
into each segment (in reverse) to retrieve the states in reverse-order. The
recursion depth bottoms out at $\log_k(T)$, at which point the algorithm has
$k$ consecutive optimizer states in memory; the algorithm then backpropagates
along this segment, before deleting all these states from memory and then
reinstantiating the next $k$-length segment of optimizer states. 
We provide additional details on the algorithm in Appendix~\ref{app:replay}.
\replay{} unlocks computing large-scale metagradients by requiring only
logarithmic storage and additional compute time.

\begin{remark}[Connection to rematerialization]
In a broad sense, both \replay{} and step-wise AD above can be viewed as
special cases of a classical approach in AD (and computing broadly) known as
\textit{rematerialization}~\citep{chaitin1981register,briggs1992rematerialization,zweig2000exact,griewank2008evaluating,chen2016training}.
To our knowledge, however, \replay{} is the first application of this 
particular rematerialization technique to the problem of computing metagradients through 
model training.
\end{remark}

\begin{remark}[Reversible learning]
  An alternative approach to calculating metagradients that does not save any 
  state is \underline{\smash{reversible}} learning~\citep{maclaurin2015gradient},
  for which one can ``invert'' previous training states from future ones.
  We focus here on general (non-reversible) learning algorithms for two reasons: first, 
  even simple algorithms such as SGD without momentum are non-reversible; second, 
  reversibility in practice introduces numerical precision issues.
\end{remark}

%% file: figs/tree.tex
\begin{forest}
    for tree={
      fit=band,
      minimum size=2.2em,
      s sep=0.46em,
    rectangle,
    fit=band,
    draw,
    calign primary child=1,
    calign=child edge
    },
    [{$0$}, simple cylinder, fill=gray!20, name=el1 %
      [{$0$}, simple cylinder, fill=gray!20
        [$\cdots$,circle,draw=none
          [{$0$}
            [{$0$}]
            [{$1$}]]
          [$\cdots$,circle,draw=none
            [$\cdots$,circle,draw=none]]]
        [$\cdots$,circle,draw=none
          [$\frac{n}{4}$
            [$\frac{n}{4}$] [$\frac{n}{4}+ 1$]]
          [$\cdots$,circle,draw=none
            [$\cdots$,circle,draw=none]]]]
      [$\frac{n}{2}$, simple cylinder, fill=gray!20, name=el2
        [$\cdots$,circle,draw=none,
          [$\frac{n}{2}$
            [$\frac{n}{2}$]
            [$\frac{n}{2}+ 1$]]
          [$\cdots$,circle,draw=none
            [$\cdots$,circle,draw=none]]]
        [$\cdots$,circle,draw=none, name=el3
          [$\frac{3n}{4}$, simple cylinder, fill=gray!20,name=el8
            [$\frac{3n}{4}$, simple cylinder, fill=gray!20]
            [$\frac{3n}{4} \footnotesize{+1}$, simple cylinder, minimum height=1em, minimum size=2em, text height=.75em, text depth=3.4pt, fill=gray!20,name=el9]]
          [$\cdots$,circle,draw=none,name=el35
            [$\cdots$,circle,draw=none,name=el5]
            [{$n\shortminus{}1$},name=el4]]]]]
    \begin{scope}[>=Triangle]
        \path [draw, <->] (current bounding box.south west) +(-20pt,0) coordinate (c) -- (c |- current bounding box.north) node [pos=.5, fill=white] {$\log_2(n)$};
    \end{scope}
    \draw[->,thick,color=blue] (el1) to[out=40,in=110] node [pos=0.40, fill=white, yshift=-4pt] {Traversal order} (el2); %
    \draw[->,thick,color=blue] (el2) to[out=40,in=110] (el3);
    \draw[->,thick,color=blue] (el3) to[out=40,in=110] (el35);
    \draw[->,thick,color=blue] (el35) to[out=east,in=north] (el4);
    \draw[->,thick,color=blue] (el4) to[out=south,in=south] (el5);
    \draw[->,thick,color=blue] (el5) to[out=north west,in=south west] (el35);
    \draw[->,thick,color=blue] (el35) to[out=west,in=south east] (el3);
    \draw[->,thick,color=blue] (el3) to[out=292,in=70] (el8);
    \draw[->,thick,color=blue] (el8) to[out=east,in=north] (el9);
    \begin{scope}[>=Triangle]
    \path [draw, <->] (current bounding box.south west) +(43.75pt,-1pt) coordinate (c) -- (current bounding box.south east) node [pos=.5, fill=white] {$n$ states}; %
    \end{scope}
    \end{forest}

%% file: sections/localpred.tex
Given a training function $f$, \replay{} enables us to compute metagradients
$\nabla f(\mathbf{z})$ for any setup $\mathbf{z}$. Can we immediately
use these metagradients to
optimize model training setups?
The answer is (generally) no:
we find that applying \replay{} to a function $f$ representing a standard model
training and evaluation routine yields metagradients that are often
$\pm\infty$-valued and generally unhelpful for optimization.
Indeed, previous work has observed similar issues optimizing
over even (very) small-scale training
\citep{bengio1994learning,pearlmutter1996investigation,maclaurin2015gradient}.

In this section, we show that an underlying source of the issue is the
{landscape} of the metaparameter optimization problem.
We then present a framework for modifying standard learning
algorithms
to admit useful metagradients, i.e., to be {\em metasmooth}.
To use a familiar analogy:
just as residual connections and improved initialization schemes can improve
optimization in standard deep learning algorithms, our framework introduces an
analogous set of modifications to enable optimization with metagradients.

\subsection{The metaparameter optimization landscape}
\label{sec:smoothness}
We first review the notion of smoothness from optimization theory,
and then adapt it to the setting of metagradients.
The resulting {\em metasmoothness} metric allows us to quantify
(and later, improve) the amenability
of the metaparameter optimization problem to gradient-based methods.

\paragraph{Smoothness.}
In optimization theory,
the basic property of a function that controls how effectively it can
be optimized
with first-order methods is {\em smoothness}.
Specifically, a function $f(\mathbf{z})$ is $\beta$-smooth at a point
$\mathbf{z}$
if its gradient $\nabla f$ satisfies the property that
\begin{equation}
  \label{eq:smoothness}
  \|\nabla f(\mathbf{z}) - \nabla f(\mathbf{z}')\| \leq \beta \cdot
  \|\mathbf{z} - \mathbf{z}'\| \qquad \text{ for all $\mathbf{z}'$},
\end{equation}
or in other words, if its gradient does not change too quickly around
$\mathbf{z}$.
To motivate this definition:
if a function $f$ is
$\beta$-smooth at $\mathbf{z}$, then a step of gradient descent with step size
$\nicefrac{1}{\beta}$ will successfully decrease the value of the function:
\begin{align*}
  f\left(\mathbf{z} - \frac{1}{\beta} \nabla f(\mathbf{z})\right)
  \leq f(\mathbf{z}) - \frac{1}{2\beta}\|\nabla f(\mathbf{z})\|^2.
\end{align*}
This guarantee makes $\beta$-smoothness
a good measure of gradient utility.

\paragraph{Metasmoothness.}
There are two main challenges in adapting the
smoothness property to the
metagradient setting.
First, evaluating \eqref{eq:smoothness} requires a search over all
possible $\mathbf{z}'$,
which is infeasible.
Second, even if we could exactly evaluate the left-hand side of
\eqref{eq:smoothness},
it would be difficult to disentangle non-smoothness of the training function $f$
from potential error in metagradient computation (e.g., a numerically unstable
operation in \replay{}).

To sidestep these issues, we propose a metric called {\em metasmoothness},
given in Definition \ref{def:meta_smoothness}.
Metasmoothness is cheap to compute---requiring only three evaluations of the
training function---and does not rely on metagradient computation.
For the remainder of this section, we fix a small constant $h > 0$,
and define the corresponding
finite-differences estimator of the directional derivative $\Delta_f$ as
\begin{align*}
  \Delta_f(\mathbf{z}; \mathbf{v}) := \frac{f(\mathbf{z} +
  h\mathbf{v}) - f(\mathbf{z})}{h}.
\end{align*}

\begin{definition}[Metasmoothness of $f$ at $\mathbf{z}$ towards $\mathbf{v}$]
  \label{def:meta_smoothness}
  Consider a training function $f$ mapping metaparameters $\mathbf{z}
  \in \mathbb{R}^n$
  to model output $f(\mathbf{z}) \in \mathbb{R}$.
  Given a metaparameter $\mathbf{z}$ and
  a vector $\mathbf{v} \in \mathbb{R}^n$,
  the metasmoothness of $f$ at $\mathbf{z}$ towards $\mathbf{v}$ is given by
  \begin{align}
    \label{eq:meta_smoothness_approx}
    S_{h, \mathbf{v}}(f; \mathbf{z})
    &\coloneqq
    \bigg|\frac{\Delta_f(\mathbf{z} + h \mathbf{v}) -
    \Delta_f(\mathbf{z})}{h} \bigg|.
  \end{align}
\end{definition}
\noindent
Definition \ref{def:meta_smoothness} measures the rate of change of
the derivative of $f(\mathbf{z})$ in the direction of a given vector
$\mathbf{v}$,
and is therefore related to $\beta$-smoothness in that:
\begin{itemize}
  \item[(a)] If $f$ is $\beta$-smooth at $\mathbf{z}$, then $S_{h,
    \mathbf{v}}(f; \mathbf{z}) \leq \beta$ for any $(h, \mathbf{v})$
    (so Definition \ref{def:meta_smoothness} is {\em necessary} for smoothness).
  \item[(b)] If $\lim_{h \to 0} S_{h, \mathbf{v}}(f; \mathbf{z}) \leq
    \beta$ for all $\mathbf{z} \in \mathbb{R}^n$ and $\mathbf{v} \in
    \mathbb{S}^{n-1}$,
    then $f$ is $\beta$-smooth everywhere
    (so a global version of Definition \ref{def:meta_smoothness} is
    {\em sufficient} for smoothness).
\end{itemize}

\paragraph{Empirical metasmoothness.}
Definition \ref{def:meta_smoothness} lets us measure the
meta-smoothness of a
training function $f$ at a particular metaparameter $\mathbf{z}$
(towards a direction $\mathbf{v}$).
This definition, however, has two shortcomings.
First, recall that the training function $f$ is a composition of a learning
algorithm $\mathcal{A}$ and an output function $\phi$,
so the smoothness of $f$ depends on that of both $\mathcal{A}$
and $\phi$ (in particular,
  $\nicefrac{\partial f}{\partial \mathbf{z}} = \nicefrac{\partial
  \phi}{\partial
\mathcal{A}}\cdot \nicefrac{\partial \mathcal{A}}{\partial \mathbf{z}}$).
Since the output
function $\phi$ might be unknown ahead of time, we are most interested in
measuring the {\em overall} metasmoothness of a learning algorithm
$\mathcal{A}$.
Second, while the result of \eqref{eq:meta_smoothness_approx} does have a
concrete basis in optimization theory, it may not be easy to
interpret in practice
(e.g., what does $S = 200$ mean?).
We address both issues simultaneously by
(a) proposing an interpretable ``binarized''
version of Definition \ref{def:meta_smoothness}, and
(b) studying metasmoothness in the space of model parameters $\theta$,
instead of the output space.
\begin{definition}[Empirical metasmoothness of $\mathcal{A}$]
  \label{def:avg_meta_smoothness}
  Let $\mathcal{A}$ be a learning algorithm which maps metaparameters
  $\mathbf{z} \in \mathbb{R}^n$
  to model parameters $\theta \in \mathbb{R}^d$,
  let $\mathbf{z}$ be a metaparameter vector,
  and let $\mathbf{v}$ be a given direction.
  Let $\mathbf{d} \in \mathbb{R}^d$ be the
  per-coordinate variation in $\theta$, i.e.,
  \begin{align*}
    \mathbf{d} = |\mathcal{A}(\mathbf{z} + 2h\mathbf{v}) -
    \mathcal{A}(\mathbf{z})|  %
  \end{align*}
  The empirical $(h, \mathbf{v})$-metasmoothness of $\mathcal{A}$ at
  $\mathbf{z}$ is given by
  \begin{align}
    \label{eq:avg_meta_smoothness}
    \widehat{S}_{h, \mathbf{v}}(\mathcal{A}; \mathbf{z})
    & =
    \mathrm{sign}(\Delta_{\mathcal{A}}(\mathbf{z};\mathbf{v}))^\top\cdot
    \mathrm{diag}\left(
      \frac{\mathbf{d}}{\|\mathbf{d}\|_1}
    \right)\cdot \mathrm{sign}(\Delta_{\mathcal{A}}(\mathbf{z} +
    h\mathbf{v};\mathbf{v})),
  \end{align}
  weights each parameter by its range.
\end{definition}
Intuitively, \eqref{eq:avg_meta_smoothness} measures the
agreement in sign between the (finite-difference approximation
of the) metagradient in the direction of $\mathbf{v}$ at $\mathbf{z}$
and at $\mathbf{z} + h\mathbf{v}$,
averaged across parameter coordinates and weighted by
the variation in each coordinate.
Taking a weighted average of sign agreements ensures that
$\smash{\widehat{S} \in [-1, 1]}$ (making it easier to interpret than
Definition \ref{def:meta_smoothness}).
The $\smash{\mathrm{diag}(\nicefrac{\mathbf{d}}{\|\mathbf{d}\|_1})}$ term
weights each agreement proportionally to the scale of the
corresponding parameter change (downweighting, e.g., coordinates $i$ that
are essentially constant).
Finally, observe that Definition \ref{def:avg_meta_smoothness} is efficient
to compute in practice: it requires only three calls to the learning
algorithm $\smash{\mathcal{A}}$.

\begin{remark}[]
  Ideally, recalling the smoothness definition \eqref{eq:smoothness},
  we would evaluate metasmoothness in all possible directions
  $\mathbf{v}$ and all points $\mathbf{z}$. Empirically, we find in
  the sequel (Section \ref{sec:improving})
  that this single-direction approximation at a single point $\mathbf{z}$
  still yields a useful estimate of metasmoothness
  (e.g., one that correlates with metagradient utility).
\end{remark}

\begin{figure}
  \centering
  \begin{subfigure}[t]{0.61\textwidth}
    \captionsetup{labelformat=empty}%
    \input{pgffigs/smoothness_vs_acc/smoothness_vs_acc.tex}
    \caption{(a)\quad\quad\quad\quad\quad\quad\quad\quad\quad\quad\quad\quad\quad}
  \end{subfigure}
  \begin{subfigure}[t]{0.37\textwidth}
    \captionsetup{labelformat=empty}%
    \raisebox{0.0135\height}{\input{pgffigs/smoothness_vs_opt/smoothness_vs_opt.tex}}
    \caption{(b)\quad\quad\quad\quad}
  \end{subfigure}
  \caption{
    \textit{(a)} For a variety of training configurations of a ResNet-9
    model, we plot metasmoothness (Def. \ref{def:avg_meta_smoothness})
    against test accuracy. Strategies such as increasing width, placing
    batch normalization before activations, and scaling down network
    outputs consistently improve  metasmoothness, at a minor cost to
    accuracy.
    \textit{(b)} Smoother training configurations can be optimized via
    metagradients more effectively. Here, as in Section
    \ref{sec:poisoning}, we use metagradients to gradient ascend on
    validation loss.
  }
  \label{fig:smoothness_vs_accuracy}
\end{figure}

\subsection{Estimating and improving metasmoothness}
\label{sec:improving}
Having established a method for quantifying metasmoothness, we turn to the
practical question: how can we design learning algorithms that are amenable to
metagradient optimization? To answer this question,
we introduce a
straightforward
framework: given a learning algorithm,
explore a fixed menu of possible modifications to the
training setup, and choose the combination that maximizes empirical
metasmoothness.
In practice, we find that this framework allows us to slightly modify learning
algorithms in a way that makes them amenable to first-order methods.

As a case study, we
study the task of training ResNet-9 on the CIFAR-10 dataset
\citep{krizhevsky2009learning}.
We let the metaparameters $\mathbf{z}$ be
a perturbation to the pixels of 1000 random training images
(so $\smash{\mathbf{z} \in \mathbb{R}^{1000 \times 32 \times 32 \times 3}}$).
We estimate
the empirical metasmoothness of different learning algorithms $\mathcal{A}$
at $\mathbf{z} = \bm{0}$ using Definition \ref{def:avg_meta_smoothness}.
Concretely, we proceed as follows for each learning algorithm $\mathcal{A}$:
\begin{enumerate}
    \item Let $\mathbf{z}_0 = \bm{0}$ be the metaparameter corresponding to
    the original dataset.
    \item Sample a random perturbation vector $\mathbf{v} \sim \mathcal{N}(0, 1)$.
    \item Compute the empirical metasmoothness \eqref{eq:avg_meta_smoothness}, i.e.,
    \begin{enumerate}
        \item Let $\theta_0 := \mathcal{A}(\mathbf{z}_0)$,
        $\theta_h := \mathcal{A}(\mathbf{z}_0 + h \cdot \mathbf{v})$, and
        $\theta_{2h} := \mathcal{A}(\mathbf{z}_0 + 2h \cdot \mathbf{v})$ be the model parameters
        that result from training with training dataset perturbations
        $\mathbf{z}_0$, $\mathbf{z}_0 + h \mathbf{v}$,
        and $\mathbf{z}_0 + 2h \mathbf{v}$, respectively.
        \item Compute the approximate derivatives
        \[
            \Delta_{\mathcal{A}}(\mathbf{z}_0; \mathbf{v}) = \left(\theta_h - \theta_0\right)/h, \quad
            \Delta_{\mathcal{A}}(\mathbf{z}_0 + h \mathbf{v}; \mathbf{v}) = \left(\theta_{2h} - \theta_h\right)/h.
        \]
        \item Compute the weighting vector $\mathbf{d}  = |\theta_{2h} - \theta_0|$, and compute the average metasmoothness \eqref{eq:avg_meta_smoothness}, i.e.,
        \[
            \widehat{S}_{h, \mathbf{v}}(\mathcal{A}; z_0) =
            \textrm{sign}(\Delta_{\mathcal{A}}(\mathbf{z}_0 + h \mathbf{v};\mathbf{v}))^\top
            \cdot \textrm{diag}\left(\frac{\mathbf{d}}{\|\mathbf{d}\|_1}\right)
            \cdot \textrm{sign}(\Delta_{\mathcal{A}}(\mathbf{z}_0;\mathbf{v})).
        \]
    \end{enumerate}
\end{enumerate}

\begin{figure}[h]
  \centering
  \newcommand{\plotnumber}{34}
  \newcommand{\zmin}{0}
  \newcommand{\zmax}{12.5}
  \newcommand{\zdist}{3}
  \input{pgffigs/smoothness_3d/smoothness_3d.tex}

  \vspace{.25cm} %

  \renewcommand{\plotnumber}{811}
  \renewcommand{\zmin}{-1}
  \renewcommand{\zmax}{3}
  \renewcommand{\zdist}{1}
  \input{pgffigs/smoothness_3d/smoothness_3d.tex}
  \caption{
    The effect of metasmoothness on the optimization landscape.
    Each plot above visualizes the loss landscape of a (deterministic)
    learning algorithm $\mathcal{A}$, with the $x$- and $y$-axes representing
    additive perturbations to 1000 examples in the training set
    and the $z$-axis representing the resulting model's loss
    on the test example given in the title.
    In each row, the left plot is a non-smooth algorithm,
    and the right plot is a smooth algorithm
    (as per Definition \ref{def:avg_meta_smoothness})
    evaluated on the same  example.
    Overall, empirical metasmoothness seems to strongly correlate with
    qualitative landscape smoothness. See Figure \ref{fig:more_landscapes}
    for more examples.
  }
  \label{fig:smoothness_3d}
\end{figure}

\paragraph{Metasmooth learning algorithms.}
We apply the procedure above to estimate the metasmoothness of learning
algorithms induced by different design choices
(batch size, network width, BatchNorm placement, gradient scaling),
and report the results in Figure \ref{fig:smoothness_vs_accuracy}
(left). On one hand,
``standard'' learning algorithms (i.e., those designed without metasmoothness in
mind) are not metasmooth.
On the other hand, our investigation reveals central factors driving
metasmoothness.
In addition to ``standard''
hyperparameters such as batch size and network width playing a role, we find
that placing Batch Normalization layers {\em prior} to nonlinearities (instead
of after) and scaling the final layer output are both crucial to metasmoothness.
Note that the modifications we consider above are not exhaustive---see
Appendix \ref{app:poisoning} for the full training setup.

Finally, in Figure~\ref{fig:smoothness_3d}, we plot
the optimization landscape of both metasmooth (right) and
non-metasmooth (left) models.
We find that the landscapes of metasmooth models are much smoother
and---qualitatively---more straightforward to optimize.

\paragraph{Metasmoothness/performance tradeoffs?}
Figure \ref{fig:smoothness_vs_accuracy} (left) relates
metasmoothness to model accuracy for the considered learning algorithms.
While there is no clear trend, the top-performing
learning algorithms are not always metasmooth.
However, the trade-off is not too severe: the most metasmooth
algorithms still achieve near-optimal accuracy. Furthermore, it is possible
that with additional searching we could identify even more accurate metasmooth
models.
Taken together with our previous experiment, our results suggest that
jointly searching
over metasmoothness and model accuracy is a general recipe for designing
learning algorithms that are both performant and metasmooth.
Finally, as we discuss in Section \ref{sec:discussion},
a fruitful avenue for future work may be to
design metasmooth learning algorithms directly,
i.e., without relying on stability heuristics or grid search.

\paragraph{Does metasmoothness aid downstream optimization?}
Recall that our motivation for studying metasmoothness is to develop learning
algorithms that we can optimize the metaparameters of via metagradients (using
first-order methods). We started with the notion of $\beta$-smoothness from
optimization theory, and we adapted it to the setting of metagradients by making
a series of approximations and modifications. The final question we address is:
does our final notion of metasmoothness actually predict the utility of
metagradients for optimization? Figure~\ref{fig:smoothness_vs_accuracy}
(right) demonstrates
that metasmoothness strongly predicts our ability to optimize the
metaparameters of a
given learning algorithm. We use metagradients (computed by
\replay{}) to gradient
ascend on validation loss with respect to the metaparameters $\mathbf{z}$, and
measure the change in model loss.

%% file: pgffigs/smoothness_vs_acc/smoothness_vs_acc.tex
\pgfplotsset{colormap/Set2}
\pgfplotsset{grid style={dashed,gray}}

\begin{tikzpicture}
    \begin{axis}[
      xlabel={Accuracy (\%)},
      ylabel={Smoothness},
      legend style={at={(1.1,1)}, anchor=north west, inner sep={6pt}},
      legend cell align={left},
      legend columns={3},
      height={6.8cm},
      width={6cm},
      colormap name={Set2},
      grid={both}
    ]

    \pgfkeys{
        /markarray/0/.initial=*,
        /markarray/1/.initial=triangle*,
        /markarray/2/.initial=square*,
        /markarray/3/.initial=o,
        /markarray/4/.initial=square*,
        /markarray/5/.initial=square
    } 
    \addplot [
      scatter,
      only marks,
      forget plot,
      visualization depends on=\thisrow{bs} \as \bsize,
      visualization depends on=\thisrow{final_scale} \as \fscale,
      visualization depends on=\thisrow{batchnorm_before_act} \as \bnact,
      visualization depends on=\thisrow{width_multiplier} \as \width,
      scatter/@pre marker code/.code={
        \pgfmathparse{\width == 1.0 ? 0 : (\width == 1.5 ? 1 : 2)}
        \edef\currentmark{\pgfkeysvalueof{/markarray/\pgfmathresult}}
        \pgfmathsetmacro{\mycolor}{\fscale == 0.125 ? 0 : (\fscale == 0.5 ? 128 : 255)}
        \pgfmathsetmacro{\myopacity}{\bnact == 0 ? 0.2 : 1}
        \pgfplotsset{mark=\currentmark}
        \def\markopts{mark size={(\bsize == 250 ? 2 : (\bsize == 500 ? 3 : 4))},%
        color of colormap=\mycolor,
        draw=black,
        opacity=\myopacity}
        \expandafter\scope\expandafter[\markopts]
      },
      scatter/@post marker code/.code={
        \endscope
      }, 
    ]
    table[
      x expr={\thisrow{avg_acc}*100},
      y=smoothness,
      col sep=comma,
    ]{pgffigs/smoothness_vs_acc/smoothness_vs_acc.csv};
    
    \addlegendimage{only marks, mark=none, draw=none, mark options={fill = white, draw = white}}%
    \addlegendentry{\makebox[0pt][l]{\textbf{Final scale}}}
    \addlegendimage{empty legend}
    \addlegendentry{}
    \addlegendimage{empty legend}
    \addlegendentry{}
    \addlegendimage{only marks, mark=*, mark options={color of colormap=0}}%
    \addlegendentry{0.125}
    \addlegendimage{only marks, mark=*, mark options={color of colormap=128}}%
    \addlegendentry{0.5}
    \addlegendimage{only marks, mark=*, mark options={color of colormap=255}}%
    \addlegendentry{1.0}

    \addlegendimage{only marks, mark=none, draw=none, mark options={fill = white, draw = white}}%
    \addlegendentry{$\phantom{\bigg|}$\makebox[0pt][l]{\textbf{Batch size}}}
    \addlegendimage{empty legend}
    \addlegendentry{}
    \addlegendimage{empty legend}
    \addlegendentry{}
    \addlegendimage{only marks, mark=*, mark size=2, mark options={fill=black}}%
    \addlegendentry{250}
    \addlegendimage{only marks, mark=*, mark size=3, mark options={fill=black}}%
    \addlegendentry{500}
    \addlegendimage{only marks, mark=*, mark size=4, mark options={fill=black}}%
    \addlegendentry{1000}

    \addlegendimage{only marks, mark=none, draw=none, mark options={fill = white, draw = white}}%
    \addlegendentry{$\phantom{\bigg|}$\makebox[0pt][l]{\textbf{Width}}}
    \addlegendimage{empty legend}
    \addlegendentry{}
    \addlegendimage{empty legend}
    \addlegendentry{}
    \addlegendimage{only marks, mark=*, mark size=3, mark options={fill=black}}%
    \addlegendentry{1.0}
    \addlegendimage{only marks, mark=triangle*, mark size=3, mark options={fill=black}}%
    \addlegendentry{1.5}
    \addlegendimage{only marks, mark=square*, mark size=3, mark options={fill=black}}%
    \addlegendentry{2.0}

    \addlegendimage{only marks, mark=none, draw=none, mark options={fill = white, draw = white}}%
    \addlegendentry{$\phantom{\bigg|}$\makebox[0pt][l]{\textbf{BN Placement}}}
    \addlegendimage{empty legend}
    \addlegendentry{}
    \addlegendimage{empty legend}
    \addlegendentry{}

    \addlegendimage{only marks, mark=*, mark size=3, opacity=1, mark options={fill=black}}%
    \addlegendentry{\makebox[0pt][l]{Before activation}}
    \addlegendimage{empty legend}
    \addlegendentry{}
    \addlegendimage{empty legend}
    \addlegendentry{}

    \addlegendimage{only marks, mark=*, mark size=3, opacity=0.2, mark options={fill=black}}%
    \addlegendentry{\makebox[0pt][l]{After activation}}
    \addlegendimage{empty legend}
    \addlegendentry{}
    \addlegendimage{empty legend}
    \addlegendentry{}

    \end{axis}
\end{tikzpicture}

%% file: pgffigs/smoothness_vs_opt/smoothness_vs_opt.tex
\pgfplotsset{colormap/Set2}
\pgfplotsset{scaled y ticks=false}
\pgfplotsset{grid style={dashed,gray}}

\begin{tikzpicture}
    \begin{axis}[
      xlabel={Smoothness},
	  ylabel={$\Delta$ Loss (Optimization Objective)},
      height={6.8cm},
      width={6cm},
	  ytick pos=right,
	  y tick label style={
        /pgf/number format/fixed,     %
        /pgf/number format/precision=2,  
        /pgf/number format/zerofill   %
      },
      scaled y ticks=false,
      colormap name={Set2},
      grid={both}
    ]

    \pgfkeys{
        /markarray/0/.initial=*,
        /markarray/1/.initial=triangle*,
        /markarray/2/.initial=square*,
        /markarray/3/.initial=o,
        /markarray/4/.initial=square*,
        /markarray/5/.initial=square
    } 
    \addplot [
      scatter,
      only marks,
      forget plot,
      visualization depends on=\thisrow{bs} \as \bsize,
      visualization depends on=\thisrow{final_scale} \as \fscale,
      visualization depends on=\thisrow{batchnorm_before_act} \as \bnact,
      visualization depends on=\thisrow{width_multiplier} \as \width,
      scatter/@pre marker code/.code={
        \pgfmathparse{\width == 1.0 ? 0 : (\width == 1.5 ? 1 : 2)}
        \edef\currentmark{\pgfkeysvalueof{/markarray/\pgfmathresult}}
        \pgfmathsetmacro{\mycolor}{\fscale == 0.125 ? 0 : (\fscale == 0.5 ? 128 : 255)}
        \pgfmathsetmacro{\myopacity}{\bnact == 0 ? 0.2 : 1}
        \pgfplotsset{mark=\currentmark}
        \def\markopts{mark size={(\bsize == 250 ? 2 : (\bsize == 500 ? 3 : 4))},%
        color of colormap=\mycolor,
        draw=black,
        opacity=\myopacity}
        \expandafter\scope\expandafter[\markopts]
      },
      scatter/@post marker code/.code={
        \endscope
      }, 
    ]
    table[
      x={smoothness},
      y={val_acc_x},
      col sep=comma,
    ]{pgffigs/smoothness_vs_opt/smoothness_vs_opt.csv};
    \end{axis}
\end{tikzpicture}

%% file: pgffigs/smoothness_3d/smoothness_3d.tex
\begin{tikzpicture}
    \centering
    \pgfplotsset{
        view={45}{30},
        mesh/ordering=y varies,
        table/col sep=comma,
        table/header=true,
    }

    \pgfplotsset{
        colormap={crest}{
            rgb=(0.647,0.803,0.566) %
            rgb=(0.500,0.737,0.568) %
            rgb=(0.357,0.651,0.565) %
            rgb=(0.247,0.586,0.562) %
            rgb=(0.154,0.471,0.548) %
            rgb=(0.113,0.355,0.518) %
            rgb=(0.132,0.289,0.486) %
            rgb=(0.173,0.190,0.445) %
        }
    }

    \pgfplotstableread[col sep=comma,header=true]{pgffigs/smoothness_3d/data/nonsmooth_3dplot_\plotnumber.csv}\nonsmoothdata
    \pgfplotstableread[col sep=comma,header=true]{pgffigs/smoothness_3d/data/smooth_3dplot_\plotnumber.csv}\smoothdata

    \begin{groupplot}[
        group style={
            group size=2 by 1,
            horizontal sep=0.1\textwidth,
            vertical sep=0.1\textwidth,
        },
        width=0.45\textwidth,
        height=0.35\textwidth,
        colormap name=viridis,
        xlabel=$h_1$,
        ylabel=$h_2$,
        zlabel=$f(h_1\mathbf{v}_1 + h_2\mathbf{v}_2)$,
        xlabel shift=-10pt,
        ylabel shift=-10pt,
        zlabel shift=-5pt,
        title style={yshift=-5pt},
        scaled ticks=true,
        xmin=0.001,
        ymin=0.001,
        zmin=\zmin,
        zmax=\zmax,
        grid=both,
        grid style={line width=0.5pt, draw=gray!30, dashed},
        xtick distance=0.25,
        ytick distance=0.25,
        ztick distance=\zdist,
        extra x ticks={0},
        extra y ticks={0},
        extra z ticks={0},
        extra tick style={grid=none, tick style={draw=none}, major tick length=0pt},
        extra x tick labels={},
        extra y tick labels={},
        extra z tick labels={}
    ]
        \nextgroupplot[title={Non-smooth (Example \#\plotnumber)}]
            \addplot3[
                surf,
                shader=interp,
                mesh/rows=21,
                mesh/cols=21
            ] table[x=X, y=Y, z=Z] {\nonsmoothdata};
            
        \nextgroupplot[title={Smooth (Example \#\plotnumber)}]
            \addplot3[
                surf,
                shader=interp,
                mesh/rows=21,
                mesh/cols=21
            ] table[x=X, y=Y, z=Z] {\smoothdata};
    \end{groupplot}
\end{tikzpicture}

%% file: sections/applications.tex
In this section, apply metagradients to three problems in
machine learning: selecting training data, poisoning training data, and
searching for hyperparameters. In each setting we follow the same recipe:
we frame the task as an optimization problem, modify the learning algorithm
of interest to be \textit{smooth}, then solve by first-order
optimizing with meta-gradients---which we refer to, in a catch-all manner across
algorithms, as metagradient descent (MGD). In particular:
we substantially improve on existing dataset selection methods
(Section \ref{sec:datacomp},
Section \ref{sec:less}), perform the first effective
accuracy-degrading data poisoning attack
(Section \ref{sec:poisoning}), and discover one-cycle learning rate
schedules with MGD
(Section \ref{sec:lrsched}).

\subsection{Selecting multimodal training data}
\label{sec:datacomp}
\input{sections/applications_clip.tex}

\subsection{Selecting instruction-tuning data}
\label{sec:less}
In our second application, we select training data for instruction fine-tuning
(IFT) using the same MGD-based method detailed in Algorithm~\ref{alg:depsing} of
Section~\ref{sec:datacomp}. As with multimodal data, training on the ``right''
post-training data (such as the ``right'' IFT data) can greatly impact
deployment-time model
performance~\citep{liu2024deepseek,dubey2024llama,taori2023stanford}. MGD
improves over baselines at choosing IFT data for
MMLU~\citep{hendrycks2021measuring}, a general knowledge task, and
BBH~\citep{suzgun2022challenging}, a reasoning/chain-of-thought task.

To overview this section: we start by detailing the setting, then describe the
specifics of our MGD instantiation before concluding with results.

\begin{figure}[t]
  \centering
  \begin{subfigure}[t]{0.49\textwidth}
    \centering
    \raisebox{-0.605\height}{ \input{pgffigs/less_result/less_result.tex} }
  \end{subfigure}
  \hfill
  \begin{subfigure}[t]{0.5\textwidth}
    \centering

\input{tables/replay_vs_less_table.tex}
  \end{subfigure}
  \caption{MGD dataset selection outperforms baselines. Comparing to training
    on all the data: it achieves over double the margin of improvement of LESS
    on MMLU, and improves by $+1.5\%$ on BBH (where LESS does not improve at
  all). The $\Delta$ column denotes improvement over not filtering.}
  \label{fig:ift_dss}
\end{figure}

\paragraph{Setting.}
We adopt the setting of LESS~\citep{xia2024less}. Here, the goal is to select a
training data subset from four combined IFT datasets (Flan
  V2~\citep{longpre2023flan}, CoT~\citep{wei2022chain},
  DOLLY~\citep{conover2023free}, and Open Assistant 1
\citep{kopf2024openassistant}) to maximize accuracy on a given target task. We
consider two target tasks from LESS: MMLU (which comprises multiple choice
questions spanning a variety of disciplines) and BBH (a 23 task subset of
BIG-Bench~\citep{srivastava2022beyond}).
In this setup, the data selector can access samples from
each task built from the in-context learning prompts.
Following \citet{xia2024less}, we
fine-tune a $128$-width LoRA~\citep{huang2020dynamics} (in our work, on
Gemma-2B~\citep{team2024gemma}). See Appendix~\ref{app:less} for full details on
the tasks and learning algorithm.

\paragraph{Method.}
We split up the available task samples into two sets---a ``target'' set and a
``validation'' set---then select data with MGD (via Algorithm~\ref{alg:depsing})
by minimizing causal language modeling loss on the ``target'' set of samples.
We select hyperparameters like step size and number of SGD iterations with the
validation set; see Appendix~\ref{app:less} for more details.

\paragraph{Results.}
Comparing with two baselines---training on \textit{all} the data and training
with data selected with LESS~\citep{xia2024less}---MGD yields strictly better
training dataset selections for each target task (cf. Figure~\ref{fig:ift_dss}).
MGD improves most on BBH, a reasoning task, compared to the best
baseline (+1.5\% accuracy). On MMLU, a knowledge-based task, we
outperform baselines by slightly less compared to the best baseline (+0.8\%);
one explanation is that selecting IFT data lends more control over reasoning
than over intrinsic knowledge available in the LM.

Beyond raw accuracy, we inspect losses across each step of the optimization
process. Overall, our method improves validation loss over MGD steps
(cf. Appendix Figures~\ref{fig:bbh_mmlu_over_time}), but also
exhibits signs of overfitting. Given intuition from
overparameterized learning, we might expect this behavior: we optimize a total
of 270,679 ``weights''---each corresponding to a count for a datapoint---to
minimize loss on only a handful of test samples (cf.
Table~\ref{tab:datasets}).

\subsection{Accuracy-degrading (Huber) data poisoning}
\label{sec:poisoning}
The goal of an accuracy-degrading {\em data poisoning} attack is to degrade the
performance of a machine learning model by corrupting a small fraction of its
training data.
Here, the considered threat model is as follows. The attacker is given
a training set $\mathbf{X} = \{x_1, \ldots, x_n\}$
drawn from a distribution $P$,
and a function $\theta(\cdot)$ mapping training data to model
parameters (representing the learning algorithm used by the victim).
The attacker's goal is to return a new training set $\mathbf{X}'$ that differs
from $\mathbf{X}$ in at most $\varepsilon\cdot n$ datapoints while
inducing model
parameters $\theta(\mathbf{X}')$ that perform as poorly as possible on a freshly
drawn test set $T$ from $P$.

Formally, the adversary aims to solve the following optimization problem:
\begin{equation}
  \label{eq:poisoning}
  \arg\max_{\widetilde{x}_1, \ldots, \widetilde{x}_{n_p}}
  \mathbb{E}_{x \sim P}[\ell(x; \theta(\mathbf{X}'))],
\end{equation}
where $\mathbf{X}' = \{\widetilde{x}_1, \ldots, \widetilde{x}_{n_p},
x_{n_p+1}, \ldots, x_n\}$ and $n_p = \lfloor \varepsilon n \rfloor$.
Note that our goal is to degrade the {\em overall} model performance
on a test set $\mathbf{X}_{test}$ drawn from $P$ (in particular,
the test set $\mathbf{X}_{test}$ is {\em unknown} to the adversary).
In this way, this setting resembles
the Huber contamination model in statistics \citep{huber1964robust},
and is strictly more challenging than the usual data poisoning
settings in deep learning (e.g., backdoor attacks
  \citep{gu2017badnets} or attacks that target specific test examples
\citep{koh2017understanding}).

For large-scale machine learning models, finding strong adversaries has proven
challenging---standard loss-minimizing learning algorithms seem quite robust to
maliciously-inserted data~\citep{lu2023exploring}.
In fact, the first non-trivial accuracy degradation data poisoning attacks on
deep models were pioneered by \citet{lu2022indiscriminate} and later improved 
upon by the same set of authors \citep{lu2023exploring}.
Broadly speaking, even constructing attacks that degrade the overall performance of a
learning algorithm by more than the adversarial budget $\varepsilon$
has proven challenging.

\subsubsection{Setup}
We observe that \eqref{eq:poisoning} is a continuous optimization
problem to which we can directly apply our metagradient framework,
approximating the expectation over $P$ by a finite-sample average
over a validation set $\mathbf{X}_{val}$.
In particular, given a (randomly shuffled) training set $\mathbf{X}$
and validation set $\mathbf{X}_{val}$, we set up the
following metaparameter optimization problem (see
Section~\ref{subsec:meta_gradient}):
\begin{itemize}
  \item[(a)] the metaparameter $\mathbf{z} \in \mathcal{X}^{n_p}$ is
    a tensor of $n_p = \lfloor \varepsilon n \rfloor$
    poisoned samples;
  \item[(b)] the algorithm $\mathcal{A}$ maps metaparameters
    $\mathbf{z}$ to a trained model $\mathcal{A}(\mathbf{z})$ by
    replacing the first\footnote{In principle,
      the adversary can also decide {\em which} samples to poison, but
    for simplicity we consider this ``fixed'' case.} $n_p$ samples in
    $\mathbf{X}$ with the samples in $\mathbf{z}$
    and then training on the resulting dataset;
  \item[(c)] the output function $\phi$ evaluates average loss on the
    validation set $\mathbf{X}_{val}$.
\end{itemize}

\subsubsection{Algorithm}
To apply our first-order methods to this problem, we start by
initializing the poisoned data
to be exactly the first $n_p$ samples in $\mathbf{X}$,
\(
  \mathbf{z}^{(0)} := \{\widetilde{x}_i^{(0)} = x_i: i \in [n_p]\}.
\)
Then, for $t = 1, \ldots, T$, we sample a minibatch
$\mathbf{X}_{val}^{(t)}$ from $\mathbf{X}_{val}$
and use \replay{} to compute the metagradient
\[
  \mathbf{g}_t = \frac{d}{d\mathbf{z}} \left(
    \sum_{x \in \mathbf{X}_{val}^{(t)}} \ell(x; \mathcal{A}(\mathbf{z}^{(t-1)}))
  \right),
\]
and update the poisoned data using (projected) gradient ascent:
\[
  \mathbf{z}^{(t)} = \Pi_{\mathcal{X}}\left(
    \mathbf{z}^{(t-1)} + \eta\cdot \text{sign}(\mathbf{g}_t)
  \right),
\]
where $\Pi_{\mathcal{X}}$ is the projection operator onto the sample
space $\mathcal{X}$.
(For example, when $\mathcal{X}$ is the space of image-label pairs,
  $\Pi_{\mathcal{X}}$ clips images' pixel values to $[0, 1]$
and ensures labels are valid probability distributions.)

\begin{figure}[tp]
  \centering
  \includegraphics[width=\textwidth]{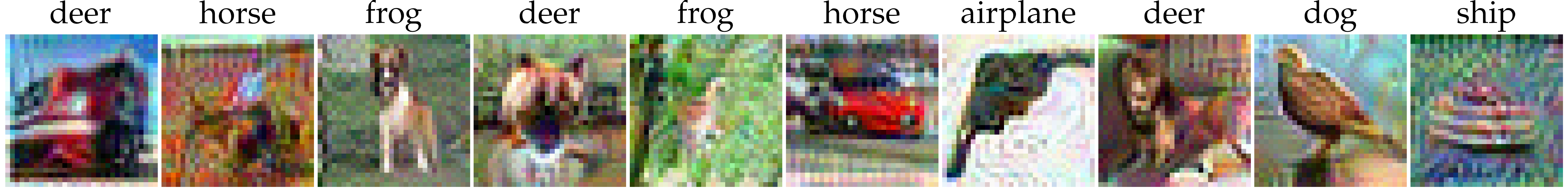}
  \caption{Examples of poisoned images from Section \ref{sec:poisoning}.}
  \label{fig:poisons}
\end{figure}

\begin{figure}[tp]
  \centering
  \begin{subfigure}[t]{0.48\textwidth}
    \centering
    \raisebox{-\height}{\input{pgffigs/cifar_poisoning/cifar_poisoning.tex}}
  \end{subfigure}
  \hfill
  \begin{subfigure}[t]{0.51\textwidth}
    \centering
    \centering
    \begin{tabular}[t]{@{}lcc@{}}\toprule
      {\bf Model} & {\bf Acc.} & $\Delta$ \\
      \midrule
      \tikz[baseline=-0.5ex] {\draw [dashed, color=pgfcolor3, very
      thick] (0,0) -- (0.6,0)} Original model & 92.0\% & $-$ \\
      \tikz[baseline=-0.5ex] {\draw [dashed, color=pgfcolor1, very
      thick] (0,0) -- (0.6,0)} GradCancel \citep{lu2023exploring} &
      91.2\% & \textcolor{mygreen}{$-$0.80\%} \\
      \tikz[baseline=-0.5ex] {\draw [color=pgfcolor0, very thick]
      (0,0) -- (0.6,0)} \textbf{MGD-DP (ours)} & \textbf{78.1\%} &
      \boldmath{\textbf{\textcolor{mygreen}{$-$13.9\%}}} \\ \midrule
      {\color{darkgray!75} 1-layer NN (for reference)
      \citep{coates2011analysis}} &
      {\color{darkgray!75} 83.3\%} & \textcolor{mygreen}{$-8.7$\%} \\
      \bottomrule
    \end{tabular}
    \label{tab:poisoning}
  \end{subfigure}
  \caption{
    For each iteration of MGD ($x$-axis),
    we train a new model from random
    initialization on a randomly shuffled training set with
    the current iterate of poisoned data injected.
    We evaluate the test accuracy ($y$-axis), and use \replay{} to compute
    the metagradient.
    MGD outperforms the best known attack \citep{lu2023exploring}
    by an order of magnitude and
    (for reference) results in a model that has the same accuracy as a
    single-layer neural network trained on random image features
    \citep{coates2011analysis}.
  }
  \label{fig:poisoning-results}
\end{figure}

\subsubsection{Evaluation}
We use the CIFAR-10 dataset
which consists of 60,000 total images each labeled as one of 10 classes.
We partition the data into 40,000 training examples, 10,000 validation examples,
and 10,000 test examples.
We consider a simple 12-epoch CIFAR-10 training procedure,
which reaches 92.4\% accuracy on the CIFAR-10 test set when applied to the
40,000 training examples. See Appendix
\ref{app:poisoning} for training hyperparameters.

As described above,
we allow the adversary to modify (in-place) a fixed, $\varepsilon$-fraction of
the training data (in our case, 2.5\%)
subject to the constraint that the poisoned images
still lay in the valid (normalized) image range of $[0, 1]$.
We compare our approach---direct optimization of the data poisoning
objective using metagradients---to the state-of-the-art ``Gradient Cancelling''
(GradCancel) method of \citet{lu2023exploring}. In short, GradCancel
is a two-step method which
first finds a poorly performing model,
then finds poisoned data that induces this model as a minimizer of
the training loss.
We present the full method in Appendix \ref{app:poisoning}.

\paragraph{Results.} We find that metagradients enable state-of-the-art
data poisoning attacks, degrading accuracy by 14\%.
In particular, when allowed to corrupt 1000 of the 40,000 training
samples (2.5\%), our method reduces test set accuracy to 78\%---for reference,
the accuracy of a single-layer neural networked trained on the unmodified
CIFAR-10 training set is 83\%.
The strongest existing data poisoning attack,
GradCancel, only reduces test set accuracy by less than 1\%.\footnote{
  \citet{lu2023exploring} report a larger drop; the discrepancy
  is due to our constraint that poisoned data are valid bounded RGB images.
}
In Figure \ref{fig:poisons}, we visualize the
poisoned images and labels found by our method.
In Figure~\ref{fig:poisoning-results},
we visualize the minibatch loss at each step of the
optimization process.

\begin{remark}[Poisoning non-smooth learning algorithms]
  Recall that to apply metagradient descent,
  we alter the learning algorithm $\mathcal{A}$
  to be metasmooth (see Section \ref{sec:smoothness}).
  This involves making modifications such as switching out
  max pooling layers for average pooling layers,
  moving batch normalization layers before activations,
  and scaling down the last layer's output by a factor of 10.
  It is natural to ask: how much does the efficacy of our method
  depend on this smoothness?
  After all, in practice the adversary cannot control the learning algorithm.
  To answer this question, we take the poison samples generated by
  MGD and insert them into the training set of a corresponding
  \smash{\underline{standard}} (i.e., non-metasmooth) learning algorithm.
  We find that our method still
  significantly degrades the performance of the model,
  from $92.8\%$ to $82.6\%$ (a drop of $10.2\%$).
\end{remark}

\subsection{Finding a learning rate schedule}
\label{sec:lrsched}
\input{sections/lr_optimization}

%% file: sections/applications_clip.tex
Curating a training dataset from a mass of unfiltered data is a necessary and
influential step in any large-scale machine learning pipeline. Deciding how to
curate such a dataset is a challenging problem that has attracted substantial
recent
interest~\citep{fang2022data,abbas2023semdedup,engstrom2024dsdm,gadre2024datacomp}.
In this section, we frame pre-training data selection as an optimization
problem, and then solve this problem by first-order optimizing with
metagradients. Applying our method to the DataComp-small benchmark
\citep{gadre2024datacomp}, we greatly improve on the state-of-the-art (our
  improvement over state-of-the-art is roughly the same as the
improvement of state-of-the-art over training on random data).

\subsubsection{Setup}
The goal of {dataset selection} is to choose a training data subset (out of
a broad pool of data) that maximizes trained machine learning model
performance.
Given this goal,
dataset selection has a natural interpretation as a
combinatorial metaparameter optimization problem.
In particular, in the language of Section \ref{subsec:meta_gradient},
for a training set of size $n$, let
\begin{itemize}
  \item[(a)] the metaparameters $\mathbf{c} \in \mathcal{C} :=
    \mathbb{Z}_{\geq 0}^n$ be non-negative data counts representing
    the number of times each training sample repeats in the training data;
  \item[(b)] the algorithm $\mathcal{A}$ be a standard large-scale
    learning procedure, which runs on a training set
    comprising $c_i$ copies of each sample $i$ for $i \in [n]$;
  \item[(c)] the output function $\phi$ be the loss of the trained model on a target
    distribution $\dtarg{}$.
\end{itemize}
Then, defining $f(\mathbf{c}) := \phi(\mathcal{A}(\mathbf{c}))$ (as
in Section~\ref{subsec:meta_gradient}),
our goal is to find the data counts $\mathbf{c^*}$ that solve
\begin{equation}
  \label{eq:ds}
  \mathbf{c}^* := \argmin_{\mathbf{c} \in \mathcal{C}} f(\mathbf{c}).
\end{equation}

\subsubsection{Gradient descent on training data}
\label{sec:gdod}
Metagradients let us \textit{directly} minimize the target task
loss \eqref{eq:ds}
with respect to the choice of training data.
At a high level, our algorithm operates as follows:
we start with a randomly chosen set of training data, then
iteratively update the dataset selection using
metagradients with respect to importance weights
placed on each training datapoint.
The specifics of our method are in Algorithm~\ref{alg:depsing};
we describe its core ideas below.

\paragraph{Idea 1: A surrogate algorithm.} We cannot use metagradients to
optimize \eqref{eq:ds} directly, because the metaparameters of
interest $\mathbf{c}$ are
discrete counts (and so the algorithm $\mathcal{A}$ is non-differentiable with 
respect to $\mathbf{c}$). 
To circumvent this problem, we relax $\mathcal{A}$: we define a surrogate algorithm
$\mathcal{A}_\mathbf{c}'$ that takes in a {\em continuous}
metaparameter $\mathbf{z} \in
\mathbb{R}^n$, whose metagradient we \textit{can} compute, then optimize using
the metagradient on $\mathcal{A}_\mathbf{c}'$.

This surrogate learning algorithm $\mathcal{A}_\mathbf{c}'$ maps a metaparameter 
$\mathbf{z} \in \mathbb{R}^n$ (representing a perturbation to training data weights) 
to a machine learning model.
The surrogate is defined by a set
of counts $\mathbf{c} \in \mathbb{Z}_+^n$, and a hyperparameter $k$ denoting a
specific training iteration, both of which we bake into the surrogate algorithm
itself. 
Given a metaparameter $\mathbf{z} \in \mathbb{R}^n$, the algorithm
$\mathcal{A}_\mathbf{c}'$ trains a model ``as usual'' using the fixed counts
$\mathbf{c}$. That is, it makes $c_i$ copies of each training sample $i$,
shuffles and partitions the data into batches, and then at each iteration
minimizes the batch loss with a step---just as the original learning algorithm
$\mathcal{A}$. At iteration $k$, however, in addition to the original loss on
the $k$-th batch, the algorithm upweights {\em each} training sample $i$
according to the metaparameter $z_i$. In other words, the objective at iteration
$t$ of the surrogate algorithm $\mathcal{A}_\mathbf{c}'$ is
\begin{equation*}
  \label{eq:surrogate_objective}
  \ell'_t(\theta) :=
  \begin{cases}
    \sum_{x \in t^{\text{th}} \text{ batch}} \ell(x; \theta) &
    \!\!\text{if } t \neq k \\
    \sum_{x \in t^{\text{th}} \text{ batch}} \ell(x; \theta) +
    \sum_{i=1}^n z_i \ell(x_i; \theta) &\!\!\text{if } t = k
  \end{cases}
\end{equation*}
where $\ell(x; \theta)$ is the training loss on example $x$.

Observe that when $\mathbf{z} = \mathbf{0}_n$, the algorithm
$\mathcal{A}_\mathbf{c}'$  is identical to the standard learning algorithm
$\mathcal{A}$. And while $\mathcal{A}$ was a function of (nondifferentiable)
discrete data counts $\mathbf{c}$, $\mathcal{A}_\mathbf{c}'$ is differentiable
with respect
its input $\mathbf{z}$, and so we can compute the metagradient
\[
  \mathbf{g} := \nabla_{\mathbf{z}}
  \phi\big(\mathcal{A}_\mathbf{c}'(\mathbf{z})\big)\big\vert_{\mathbf{z}
  = \mathbf{0}_n}.
\]
Intuitively, the entries of the metagradient $\mathbf{g}$ capture the effect of
adding an infinitesimal amount of each training sample $i$ to the training data
at iteration $k$. A positive entry $g_i$ indicates that adding an infinitesimal
amount of sample $i$ to the training data would increase the loss,  and a
negative entry indicates that adding an infinitesimal amount of sample $i$ to
the training data would decrease the loss; the slot at $i$ represents the
(estimated) effect of adding a copy of sample $i$ to the training
data at every batch containing the sample.

\paragraph{Idea 2: Block coordinate descent.}
We then use the metagradient $\mathbf{g}$ to iteratively update our selected
dataset. We update data counts as
\begin{equation}
  \label{eq:update}
  \mathbf{c} \gets \mathbf{c} - \mathrm{sign}(\mathbf{g}) \odot
  \mathbf{m}, \quad \mathbf{m} \sim \mathrm{Bernoulli}(p),
\end{equation}
where $p$ is a hyperparameter controlling the fraction of sample counts to
update. This algorithm resembles a block coordinate descent algorithm
\citep{ortega2000iterative}, with the main difference being that we take signed
gradient steps with step size $1$ (projected onto non-negative integers) to
ensure that the counts remain well-defined. As a result, $p$ implicitly controls
the algorithm's step size.

Applying \eqref{eq:update} concludes a single optimization step.
By repeating this process of estimating the metagradient, updating our
counts vector, then constructing a new training dataset, we iteratively
improve the selected data.  Pseudocode for our algorithm can be found in Algorithm \ref{alg:depsing}.

\begin{algorithm}[h]
    \SetKwFor{For}{\textcolor{forcolor}{for}}{\textcolor{forcolor}{do}}{}
    \SetKw{Return}{\textcolor{forcolor}{Return}}{}
    \SetKwFunction{traversereverse}{\textcolor{fncolor}{reverse\_inorder\_traversal}}
    \DontPrintSemicolon
	\KwIn{initial data counts $\mathbf{c} \in \mathbb{Z}_{\geq 0}^n$, learning algorithm $\mathcal{A}$, output function $\phi$}
    \nonl\hspace{0pt}{\textbf{Hyperparameters:}} step size $p$, \# opt steps $T$, iteration number $k$ \;
    \For{$t \gets 1$ \color{forcolor}{\textbf{to}} \color{black}$T$}{
		$\mathbf{z} \gets \mathbf{0}_n$ \tcp{Build input to surrogate}
		$\mathbf{g} \gets \frac{\partial \phi(\mathcal{A}'_\mathbf{c}(\mathbf{z}))}{\partial \mathbf{z}}$ \tcp{Calculate metagradient using \textsc{Replay}}
		$\mathbf{m} \gets$ sample from $\textrm{Bernoulli}(p)$ \tcp{Sample indices to step on}
		$\mathbf{c} \gets \mathbf{c} - \mathrm{sign}(\mathbf{g}) \odot \mathbf{m}$ \tcp{Take optimization step}
    }
	\Return $\mathbf{c}$ \tcp{Return final data counts}
	\caption{Dataset selection using using metagradient descent (MGD).}
    \label{alg:depsing}
\end{algorithm}

\subsubsection{Results}
We evaluate our data selection algorithm using DataComp
\cite{gadre2024datacomp}, a standardized framework for evaluating data selection
methods for multimodal models. Algorithm~\ref{alg:depsing} greatly
improves on the state-of-the-art for the benchmark. Below, we
describe the setting, outline our
method, and conclude with our results.

\paragraph{Setting.}
DataComp \citep{gadre2024datacomp} is a multimodal model training competition 
and benchmark for evaluating dataset selection methods. 
DataComp provides a \textit{fixed} learning algorithm
chosen in advance by the organizers and a large fixed \emph{candidate pool} of
internet data. The goal is to choose a subset of the
candidate pool---possibly with repeated datapoints---that yields the best-performing
model after training with the given learning algorithm, as measured by a
predetermined set of 38 benchmarks. Given a submission subset, the mean score on
the evaluation datasets for a model trained with that subset is taken as the
final ``score.''  DataComp offers four separate ``scales'' requiring different
amounts of compute; we focus on the \emph{small} scale in this paper
due to compute limitations.

\paragraph{Method.}
We select data with MGD (Algorithm~\ref{alg:depsing}) to minimize loss on
data on a ``target set'' that is distributionally similar to the DataComp
benchmark tasks, and select hyperparameters with a held-out ``validation
set.'' In particular, we construct target and validation sets by taking
samples from the DataComp evaluation tasks with extra samples available beyond
those used in the DataComp test set (e.g., ImageNet, one of the tasks in
DataComp, has a training set in addition to the test set evaluated in DataComp).
See Appendix~\ref{app:clip_details} for the exact details of the target and
validation sets, the precise hyperparameters used with
Algorithm~\ref{alg:depsing}, and a discussion on scalability (including further
engineering details on executing our algorithm efficiently).

\paragraph{Results.}
MGD greatly outperforms the current state-of-the-art: the difference in accuracy
between MGD and the current best method is roughly as large as the difference
between the previous state-of-the-art (EcoDatum~\citep{ecodatum2024ecodatum})
and training on randomly chosen data
(cf.\ Figure~\ref{fig:datacomp_score_over_time}).
Inspecting scores over the course of the optimization in Figure
\ref{fig:datacomp_score_over_time}, we find that only a few steps are necessary
to outperform previous methods.

\begin{figure}[ht]
  \centering
  \begin{subfigure}[t]{0.49\textwidth}
    \raisebox{-\height}{\input{pgffigs/datacomp_small_result/datacomp_small_result.tex}}
  \end{subfigure}
  \hfill
  \begin{subfigure}[t]{0.5\textwidth}
    \input{tables/clip_improvement_table.tex}
  \end{subfigure}
  \caption{MGD dataset selection greatly outperforms existing methods (improving
      over the previous SOTA by as much as the previous SOTA improves over no
    filtering at all). We compare DataComp scores for MGD (over
    optimization steps),
    training on the entire candidate pool, the best baseline
    originally proposed by
  DataComp, and the previous SOTA~\citep{ecodatum2024ecodatum}.}
  \label{fig:datacomp_score_over_time}
\end{figure}

%% file: pgffigs/datacomp_small_result/datacomp_small_result.tex
\pgfplotsset{colormap/Set2}
\pgfplotsset{scaled y ticks=false}
\pgfplotsset{grid style={dashed,gray}}

\begin{tikzpicture}
    \begin{axis}[
      xlabel={Metagradient steps},
      ylabel={DataComp Score},
      height={4.8cm},
      width={7.5cm},
      colormap name={Set2},
      ytick distance=0.03,
      xmin=-3,
      xmax=57,
      grid=both
    ]

    \addplot[mark=none, index of colormap=3, domain=-10:60, samples=2, dashed, very thick] {0.1377730386384371};
    \addlegendentry{No filtering}
    \addplot[mark=none, index of colormap=1, domain=-10:60, samples=2, dashed, very thick] {0.173};
    \addlegendentry{Best DataComp baseline}
    \addplot[mark=none,, index of colormap=2, domain=-10:60, samples=2, dashed, very thick] {0.182};
    \addlegendentry{Prev. SOTA (EcoDatum)}
    \addplot [scatter,
                index of colormap=0,
                very thick,
                scatter/@pre marker code/.code={3
                    \def\markopts{index of colormap=0}
                    \expandafter\scope\expandafter[\markopts]
                },
                scatter/@post marker code/.code={
                    \endscope
                },] 
    table[
      x=it,
      y={score},
      col sep=comma,
    ]{pgffigs/datacomp_small_result/datacomp_small_result.csv};
    \addlegendentry{MGD-DS (ours)}

    \legend{}
    \end{axis}
\end{tikzpicture}

%% file: tables/clip_improvement_table.tex
\begin{tabular}[t]{@{}lcc@{}}
\toprule
\textbf{Method}  &  \textbf{Score} & $\Delta$ \\
\midrule
\tikz[baseline=-0.5ex] {\draw [dashed, color=pgfcolor3, very thick] (0,0) -- (0.6,0)} Baseline: No filtering  & 0.13 & -- \\
\tikz[baseline=-0.5ex] {\draw [dashed, color=pgfcolor1, very thick] (0,0) -- (0.6,0)} Best baseline from \cite{gadre2024datacomp}  & 0.17 & \textcolor{mygreen}{+0.04} \\
\tikz[baseline=-0.5ex] {\draw [dashed, color=pgfcolor2, very thick] (0,0) -- (0.6,0)} Previous SOTA \cite{ecodatum2024ecodatum}  & 0.18 & \textcolor{mygreen}{+0.05} \\
\tikz[baseline=-0.5ex] {\draw [color=pgfcolor0, very thick] (0,0) -- (0.6,0); \filldraw [color=pgfcolor0] (0.3,0) circle (0.08);} \textbf{MGD-DS (ours)} & \textbf{0.22} & \textbf{\textcolor{mygreen}{+0.09}} \\
\bottomrule
\end{tabular}

%% file: pgffigs/less_result/less_result.tex
\pgfplotsset{colormap/Set2}
\pgfplotsset{scaled y ticks=false}
\pgfplotsset{grid style={dashed,gray}}

\begin{tikzpicture}
\begin{axis}[
    ybar,
    axis lines=box,
    ytick pos=left,
    axis on top=false,
    height=4.8cm,
    width=7.5cm,
    colormap name={Set2},
    ymin=-0.2,
    ymax=1.7,
    ylabel={$\Delta$ Accuracy (\%)},
    symbolic x coords={BBH,MMLU},
    xtick=data,
    xtick style={draw=none},
    enlarge x limits=0.5,
    bar width=15pt,
    grid=major,
    xmajorgrids=false,
    extra y ticks={0},
    extra y tick labels={},
    extra y tick style={
      grid=major,
      major grid style={black, solid}
    },
]

\addplot[
    fill=mapped color,
    draw=mapped color,
    point meta=0,
    index of colormap=1,
] coordinates {
    (BBH,-0.047672)
    (MMLU,0.540605)
};

\addplot[
    fill=mapped color,
    draw=mapped color,
    point meta=0,
    index of colormap=0,
] coordinates {
    (BBH,1.481074)
    (MMLU,1.297280)
};

\end{axis}
\end{tikzpicture}

%% file: tables/replay_vs_less_table.tex
\begin{tabular}{@{}l*{2}{cc}@{}}
	\toprule
	& \multicolumn{2}{c}{\textbf{BBH} \citep{suzgun2022challenging}} & \multicolumn{2}{c}{\textbf{MMLU} \citep{hendrycks2020measuring}} \\
	\cmidrule(lr){2-3} \cmidrule(l){4-5}
	& Acc. & $\Delta$ & Acc. & $\Delta$ \\
	\midrule
	All Data & 35.2\% & $-$ & 41.2\% & $-$ \\
	\textcolor{pgfcolor1}{\rule{0.26cm}{0.26cm}} LESS & 35.2\% & \textcolor{myred}{$-0.0$\%} & 41.8\% & \textcolor{mygreen}{$+0.5$\%} \\
	\textcolor{pgfcolor0}{\rule{0.26cm}{0.26cm}} \textbf{MGD-DS} & \textbf{36.7\%} & \textbf{\textcolor{mygreen}{$\mathbf{+1.5}$\%}} & \textbf{42.5\%} & \textbf{\textcolor{mygreen}{$\mathbf{+1.3}$\%}} \\
	\bottomrule
\end{tabular}

%% file: pgffigs/cifar_poisoning/cifar_poisoning.tex
\pgfplotsset{colormap/Set2}
\pgfplotsset{scaled y ticks=false}
\pgfplotsset{grid style={dashed,gray}}

\begin{tikzpicture}
    \begin{axis}[
      xlabel={Metagradient steps},
      ylabel={Test Accuracy},
      height={4.8cm},
      width={7.5cm},
      colormap name={Set2},
      ytick distance=0.03,
      y tick label style={
		/pgf/number format/.cd,
		zerofill,
		precision=2
	  },
      xmin=-15,
      xmax=800,
      grid=both
    ]

    \addplot[mark=none, index of colormap=3, domain=-10:1000, samples=2, dashed, very thick] {0.920};
    \addlegendentry{Original accuracy}
    \addplot[mark=none, index of colormap=1, domain=-10:1000, samples=2, dashed, very thick] {0.912};
    \addlegendentry{Gradient cancelling}
    \addplot [scatter, 
				mark size=0.0,
                index of colormap=0,
                very thick,
                scatter/@pre marker code/.code={3
                    \def\markopts{index of colormap=0}
                    \expandafter\scope\expandafter[\markopts]
                },
                scatter/@post marker code/.code={
                    \endscope
                },] 
    table[
      x=iteration,
      y={test_acc},
      col sep=comma,
    ]{pgffigs/cifar_poisoning/cifar_poisoning.csv};
    \addlegendentry{MGD-DS (ours)}

    \legend{}
    \end{axis}
\end{tikzpicture}

%% file: sections/lr_optimization.tex
As a final application, we optimize the learning rate schedule
of stochastic gradient descent (SGD) for training a CIFAR-10 classifier.
By following the metagradients with respect to the learning rate
at each step of training, our procedure matches grid searching
over standard learning rate schedules---despite starting with na\"ive
hyperparameters (a flat learning rate).

Unlike the other applications discussed here,
metagradients do not unlock state-of-the-art performance.
Instead, we discuss this application to illustrate the flexibility
of \replay{}, and in particular its ability to optimize metaparameters
that do not directly affect the loss landscape (i.e., that only affect
the model via the optimization trajectory).
As we discuss in Section \ref{sec:related},
approximate metagradient estimators cannot apply to these metaparameters.

\subsubsection{Setting}
To put learning rate schedule optimization into the metagradient framework,
we parameterize a schedule as a vector $\bm{\eta} \in \mathbb{R}^k$
comprising $k$ evenly-spaced keypoints,
so that the learning rate at iteration $t$ is given by
\begin{equation}
  \label{eq:lr_schedule}
  \eta(t) = \eta_{\lfloor kt/T \rfloor} + \frac{kt/T - \lfloor kt/T
  \rfloor}{\lceil kt/T \rceil - \lfloor kt/T \rfloor} (\eta_{\lceil
  kt/T \rceil} - \eta_{\lfloor kt/T \rfloor}),
\end{equation}
i.e., a linear interpolation between the keypoints.
\begin{itemize}
  \item[(a)] the metaparameter $\bm{\eta} \in \mathbb{R}^k$ is a
    vector of $k$ keypoints;
  \item[(b)] the algorithm $\mathcal{A}$ maps metaparameters
    $\bm{\eta}$ to a trained model $\mathcal{A}(\bm{\eta})$ by
    training a model for $T$ iterations with the learning rate schedule
    defined by \eqref{eq:lr_schedule};
  \item[(c)] the output function $\phi$ evaluates average loss on the
    validation set $\mathbf{X}_{val}$.
\end{itemize}

\subsubsection{Algorithm}
Following the theme of the rest of this section, we optimize the
metaparameter $\bm{\eta}$
directly using MGD. In particular, we initialize the keypoints to be
a flat learning rate schedule,
and then update the keypoints using the metagradient with respect to
the validation loss,
\begin{align*}
  \bm{\eta}^{(t+1)} = \bm{\eta}^{(t)} - \alpha \cdot
  \mathrm{sign}\left(\nabla_{\bm{\eta}}
  \phi(\mathcal{A}(\bm{\eta}^{(t)}))\right).
\end{align*}

\begin{figure}[tp]
    \centering
    \input{pgffigs/lr_opt_result/both_graph.tex}
    \caption{Target and test accuracies of MGD's learning rate schedule
      over time closely match or exceed those found by a grid search over
      hundreds of combinations of hyperparameters.  95\% confidence
    intervals are plotted for MGD's results.}
    \label{fig:mgd_vs_grid}
\end{figure}

\subsubsection{Evaluation}
We aim to select the learning rate schedule that minimizes the
expected test set loss.
To do so, we reserve 90\% of the CIFAR-10 test set as a ``validation set''
on which we select hyperparameters.
We then use the remaining 10\% as a test set. We compare the
following two approaches:
\begin{itemize}
  \item {\bf Grid search:} We construct a grid over different one cycle
    learning rate schedules, varying the peak learning rate, starting learning
    rate, ending learning rate, and peak learning rate time. In total, we
    consider over 1,000 different learning rate schedules. We use the reserved
    90\% of the test set to select the best learning rate schedule
    from the grid.
  \item {\bf Metagradient descent (MGD):} We run 50 steps of
    MGD starting from a highly suboptimal flat learning rate schedule, aiming to
    minimize loss on the reserved 90\% of the test set. We use the
    last iteration
    of MGD as our learned learning rate schedule.
\end{itemize}
We evaluate the performance of each final learning rate schedule on the held-out
10\% test set and average the results over the same set of 5 unseen
random seeds.

\paragraph{Results.}
Comparing our learned hyperparameter schedule to grid search,
as shown in Figure \ref{fig:mgd_vs_grid}, our learned schedule
using only 50 steps of MGD matches the performance of the state-of-the-art
onecycle schedule found via grid search over more than 1000 configurations.
An important caveat, however, is that these numbers are not directly comparable:
grid search can be run in parallel across many machines, while steps
of MGD must be run sequentially.

In practice, we do not advise using MGD for optimizing low-dimensional
hyperparameters, especially ones that have been thoroughly optimized by grid
search (such as CIFAR-10 learning rate schedules
\citep{smith2017super,page2018cifar,li2019exponential,jordan202494}). Still, an
interesting avenue for future work is to study the utility of MGD for optimizing
high-dimensional hyperparameters that are less well-studied, such as
per-parameter/layer learning rates/weight decays for language models, attention
hyperparameters, or gradient preconditioners.

%% file: pgffigs/lr_opt_result/both_graph.tex
\pgfplotsset{colormap/Set2}
\pgfplotsset{scaled y ticks=false}
\pgfplotsset{grid style={dashed,gray}}

\begin{tikzpicture}
    \begin{axis}[
      xlabel={Metagradient steps},
	  ylabel={Accuracy},
      height={5.2cm},
	  width={10cm},
      colormap name={Set2},
	  legend pos=south east,
      legend style={at={(1.1,1)}, anchor=north west, inner sep={6pt}},
      legend cell align={left},
	  xmin=-1,
	  xmax=51,
	  ymin=0.87,
	  ymax=0.95,
      grid=both
    ]
	\addlegendentry{Best grid target acc}
	\addplot[mark=none, index of colormap=3, domain=-2:1000, samples=2, dashed, very thick] {0.9401889};

	\addlegendentry{Best grid test acc}
	\addplot[mark=none, index of colormap=3, domain=-2:1000, samples=2, solid, very thick] {0.9266};

	\addplot[
      index of colormap=0,
	  mark size=0.0,
	  very thick,
	  dashed,
      error bars/.cd,
      y dir=both,
      y explicit,
    ] table[
      x=it,
      y=ymean,
      col sep=comma   %
	]{pgffigs/lr_opt_result/target_over_time.csv};
	\addlegendentry{MGD-LR target acc}

	\addplot[
      index of colormap=0,
	  mark size=0.0,
	  very thick,
      error bars/.cd,
      y dir=both,
      y explicit,
    ] table[
      x=it,
      y=ymean,
      col sep=comma   %
	]{pgffigs/lr_opt_result/test_over_time.csv};
	\addlegendentry{MGD-LR test acc}

	\addplot [
	  name path=upper_target,
	  draw=none,
	  forget plot
	] 
	table [
	  x=it, 
	  y expr=\thisrow{ymean} + \thisrow{yerr}, 
	  col sep=comma
	]{pgffigs/lr_opt_result/target_over_time.csv};

	\addplot [
	  name path=lower_target,
	  draw=none,
	  forget plot
	] 
	table [
	  x=it, 
	  y expr=\thisrow{ymean} - \thisrow{yerr}, 
	  col sep=comma
	]{pgffigs/lr_opt_result/target_over_time.csv};

    \addplot[
      fill opacity=0.2,
	  index of colormap=0,
    ] fill between[
      of=upper_target and lower_target,
    ];

	\addplot [
	  name path=upper_test,
	  draw=none,
	  forget plot
	] 
	table [
	  x=it, 
	  y expr=\thisrow{ymean} + \thisrow{yerr}, 
	  col sep=comma
	]{pgffigs/lr_opt_result/test_over_time.csv};

	\addplot [
	  name path=lower_test,
	  draw=none,
	  forget plot
	] 
	table [
	  x=it, 
	  y expr=\thisrow{ymean} - \thisrow{yerr}, 
	  col sep=comma
	]{pgffigs/lr_opt_result/test_over_time.csv};

    \addplot[
      fill opacity=0.2,
	  index of colormap=0,
    ] fill between[
      of=upper_test and lower_test,
    ];

    \end{axis}
\end{tikzpicture}

%% file: sections/discussion.tex
In this section, we first present the main limitations of our method and
outline future directions.

\paragraph{Limitations.} Although \replay{} is more efficient than existing
methods at computing metagradients, it is still non-trivially more expensive
than simply training a model once. The main reason is that metagradients require
making a \emph{backwards pass over a backwards pass}. This operation necessarily
requires 2-3 times the operations of a backwards pass; furthermore, our current
implementation requires \texttt{float32}/\texttt{tensorfloat32} operations.
Finally, standard training operations are often made more efficient by
specialized software (e.g., via FlashAttention~\citep{dao2022flashattention});
no such software (yet) exists for backwards-over-backwards operations. Beyond
computational issues, successfully applying metagradients requires smooth model
training. 

\paragraph{Metasmoothness: connections and future directions.}
While Section \ref{sec:localpred} describes a general procedure for finding 
metasmooth learning algorithms, an important future direction is 
to further explore and understand metasmoothness.
This includes, for example: 
(a) characterizing the relationship between metasmoothness 
and numerical stability (and potentially using techniques from the latter to
improve the former); 
(b) devising improved optimizers and/or architectures that lead directly 
to metasmooth learning algorithms (akin to skip connections or 
stable initialization in architecture design);
(c) formalizing connections between metasmoothness and other 
optimization-related phenomena in deep learning 
\citep{leclerc2020two,cohen2022gradient}. 
A related but separate direction is to explore the possibility of using 
techniques from non-smooth optimization \citep{clarke1990optimization} 
to perform metagradient descent on non-metasmooth learning algorithms.

\paragraph{Applying metagradients.}
Our methods apply to any ML task that requires
optimizing with respect to a metaparameter. 
These include: poisoning data
(generated or simply hosted on the internet) so that it cannot be trained on
without permission (i.e., by maximizing training loss with respect to the text);
selecting better training data at various stages of the model training
lifecycle; and designing better model training routines and architectures with
first-order methods. Another direction of future work lies in mitigating the
computational limitations of our algorithm. Both (a) small-scale
proxy-models~\citep{hoffmann2022training,engstrom2024dsdm} and (b) low-hanging
engineering improvements can likely make calculating metagradients much more
efficient.

%% file: sections/ext_related.tex
We overview previous work on calculating and applying meta-gradients.

\subsection{Calculating metagradients}
Previous work estimates the metagradient for large-scale models via one of
two broad families of methods: implicit differentiation and automatic (explicit)
differentiation. Note that in previous literature, synonyms for metagradient
include ``hyper-gradient'' and ``outer gradient.''

\paragraph{Implicit differentiation.}
One family of methods aims to {\em approximate} the metagradient.
To illustrate the idea behind such approaches, suppose that the
learning algorithm
$\mathcal{A}$ returns a model state $\theta$ that minimizes a strongly
convex loss function $\mathcal{L}(z, \theta)$. Here, the implicit function
theorem tells us that
\begin{align}
  \label{eq:ift}
  \nabla_z f(z) = \overbrace{
    \left(\frac{d\phi}{d\theta}\bigg|_{\theta = \mathcal{A}(z)}\right)
  }^{\substack{1 \times p \text{ gradient of output} \\ \text{wrt.
  final params}}}
  \underbrace{
    \left(
      \frac{\partial^2 \mathcal{L}(z, \theta)}{\partial
      \theta^2}\bigg|_{\theta = \mathcal{A}(z)}
    \right)^{-1}
  }_{\substack{p \times p \text{ inverse Hessian of loss}
  \\ \text{wrt. final params}}}
  \overbrace{
    \left(\frac{\partial^2 \mathcal{L}(z, \theta)}{\partial
    \theta\,\partial z}\bigg|_{\theta = \mathcal{A}(z)}\right).
  }^{
    \substack{p \times n \text{ Jacobian of loss gradient}
    \\ \text{wrt. metaparameters}}
  }
\end{align}
The form of \eqref{eq:ift} yields efficient and accurate estimators for
metagradients of models learned by minimizing a strongly convex
loss~\citep{bertrand2020implicit,bertrand2022implicit,kolter2020deep,blondel2022efficient,scieur2022curse}.
Such approaches can extend to estimate metagradients of large-scale, non-convex
learning
algorithms~\citep{bengio2000gradient,koh2017understanding,rajeswaran2019meta,finn2017model,lorraine2020optimizing,chen2020stabilizing,bae2022if},
but lose any correctness guarantees. Indeed, applying this class of methods in
large-scale settings is challenging as doing so requires (a) assuming conditions
on the learning algorithm (e.g., Hessian invertibility, continuous
differentiability) and (b) efficiently approximating the inverse Hessian (in
practice, typically at the cost of estimate accuracy). Finally, implicit
function-based approaches are fundamentally limited in that they can only
differentiate with respect to metaparameters expressed in the loss function
(e.g., these methods can differentiate with respect to the weight decay, but not
learning rate).

\paragraph{Automatic (explicit) differentiation.} Beyond implicit
differentiation approaches, there is a long line of work on directly calculating
metagradients with AD (see Section~\ref{sec:computing}). Previous work has used
AD to estimate metagradients of learning algorithms ranging from those with
convex objectives to small neural
networks~\citep{hara2019data,maclaurin2015gradient,franceschi2017forward,micaelli2021gradient,zhang2021idarts,chandra2022gradient,scieur2022curse}.
As detailed in Section~\ref{sec:computing}, the primary challenge with
(reverse-mode) AD-based approaches to meta-differentiation is storing the
intermediate products required for the backward pass. To circumvent this
challenge, previous work either (a) only considers settings that are small
enough that is possible to differentiate while requiring space that is linear in
the number of iterations (i.e., 2 layer networks on MNIST), (b) uses
forward-mode
AD~\citep{franceschi2017forward,micaelli2021gradient,chandra2022gradient} (which
  requires no extra storage at the cost of additional compute that
  scales linearly
with metaparameter dimension), (c) only \textit{approximates} the metagradient
by calculating over only a few training
steps~\citep{liu2018darts,chen2020stabilizing,finn2017model}, or uses (d) a
reversible learning algorithm~\citep{maclaurin2015gradient}. The fourth category
is a promising direction for reducing space requirements when computing
large-scale metagradients, but current approaches require (a) representing model
parameters in a fixed-precision format (which current large-scale learning
algorithms do not support) in addition to restricting the algorithm to be
reversible (e.g., SGD and standard GD do not qualify). A common thread is that
algorithms computing metagradients with AD often suffer from numerical
instability and overflow issues~\citep{micaelli2021gradient,scieur2022curse}. In
relation to previous work on AD, \replay{} (Section \ref{sec:computing})
can be seen as a strategy for choosing gradient
checkpointing~\citep{chaitin1981register,briggs1992rematerialization,zweig2000exact,griewank2008evaluating,chen2016training}
locations in the compute graph (an NP-complete task in
general~\citep{naumann2008optimal}).

\subsection{Applying metagradients}
Previous work applies metagradients to optimize training setup, including
distillation~\citep{maclaurin2015gradient,lorraine2020optimizing}, training data
selection~\citep{hara2019data,engstrom2024dsdm},
meta-learning~\citep{finn2017model,rajeswaran2019meta,hospedales2021meta},
learning rate/weight decay
selection~\citep{micaelli2021gradient,chandra2022gradient}, tuning data
augmentation~\citep{lorraine2020optimizing}, and architecture
search~\citep{maclaurin2015gradient,liu2018darts,zhang2021idarts}. Beyond
optimizing metagradients, methods in data attribution apply metagradients to
(Taylor) estimate the effect of dropping training data on model
predictions~\citep{koh2017understanding,grosse2023studying,park2023trak}. To the
Previous works either (a) calculate metagradients directly with AD (made
feasible by working in a very small-scale learning setting) or (b) estimate the
metagradient with an implicit function-based approach.

%% file: sections/conclusion.tex
In this work we add metagradients to the large-scale machine learning toolkit.
To do so, we overcome two challenges: (a) calculating metagradients at scale and
(b) modifying learning algorithms to be metasmooth---i.e., to admit
metagradients that locally predict model behavior. We then successfully
calculate and apply metagradients for large-scale models (up to 2B parameters)
to select data for CLIP pretraining and instruction fine-tuning, 
to (Huber) poison training data to decrease overall model accuracy, 
and search for high-dimensional hyperparameters (per-iteration learning rates). 
Given the successful applications of metagradients in these settings, we are
excited to see what unlocking metagradients enables in other areas of machine
learning.

%% file: sections/appmethods.tex
This appendix contains supplementary material for Section~\ref{sec:computing}.
We describe two algorithms in detail: step-wise AD, and our own algorithm
\replay{}. Refer to Section~\ref{sec:computing} for the notation used in this
appendix.

\subsection{Warmup: Step-wise AD}
\label{sec:ad}
We fully describe step-wise AD in Algorithm~\ref{alg:factored}. The algorithm
requires storing all $T$ optimizer states, but requires constant memory overhead
for each AD call (as each AD call is over a single step), making it feasible to
compute for small setups.

\begin{algorithm}
  \SetKwFor{For}{\textcolor{forcolor}{for}}{\textcolor{forcolor}{do}}{}
  \SetKw{Return}{\textcolor{forcolor}{Return}}{}
  \SetKwFunction{traversereverse}{\textcolor{fncolor}{reverse\_inorder\_traversal}}
  \DontPrintSemicolon
  \hspace{0pt} \tcp{Store each optimizer state on disk}
  $\{s_i\}_{i=0}^T \gets$ Train model via $A(z)$ \;
  \;
  \hspace{0pt} \tcp{Variables; shorthand for $\smash{\frac{\partial f(z)}{\partial z}}$ and $\smash{\frac{\partial f(z)}{\partial s_T}}$}
  $\bar{z} \gets 0$ \;
  $\bar{s}_T \gets \smash{\pfrac{g(s_T)}{s_T}}$ \qquad \tcp{One reverse-mode AD call} %
  \;
  \hspace{0pt} \tcp{Reverse-mode differentiate step-by-step}
  \For{$s_i \gets s_{T-1}$ \color{forcolor}{\textbf{to}} \color{black}$s_0$}{

    \hspace{0pt} \tcp{One reverse-mode AD call. Left: $\nabla_{s_i} f$. Right: contribution to $\nabla_{z} f$ at $i$.}
        $\bar{s}_i \gets \bar{s}_{i+1} \cdot \pfrac{h_{i}(s_i, z)}{s_i}, \qquad \bar{z}_i \gets \bar{s}_{i+1} \cdot \pfrac{h_{i}(s_i, z)}{z}$\;
        \;
        $\bar{z} \leftarrow \bar{z} + \bar{z}_i$ \qquad \tcp{Accumulate metagradient}
  }
  \;
  \Return $\bar{z}$ \;
  \caption{metagradients in $\mathcal{O}(T)$ space.}
  \label{alg:factored}
\end{algorithm}

\subsection{\replay{}}
\label{app:replay}
We now describe \replay{}, our method for calculating metagradients. For a free
parameter $k \in \mathbb{N}$, \replay{} requires storing $\mathcal{O}(k\log_k(T))$ optimizer
states and an additional $\mathcal{O}(\log_k(T))$ factor of computation. The
free parameter $k$ controls the trade-off between storage and required compute.
We fully describe \replay{} in Algorithm~\ref{alg:replay}. \replay{} modifies
Algorithm~\ref{alg:factored} by retrieving the optimizer states in reverse order
using a $k$-ary tree structure in lieu of a list of all the stored states.

\subsubsection{Lazy $k$-ary tree}
We now describe the $k$-ary tree structure that underlies \replay{}; for a
visual reference of this tree with $k=2$, see
Figure~\ref{fig:efficient_data_structure}. For ease of analysis we parameterize
the total number of states as $n=T+1$ (and therefore take $n-1$ total training steps) when
describing this data structure, and assume WLOG that $n$ is an integer power of
$k$. At a high level, traversing this tree recursively replays retraining to
recover all the optimizer states in reverse order, while deleting states that
are no longer needed. We call this tree ``lazy'' because it retrains only when
required to obtain states that are not yet retrieved.

The tree is a complete $k$-ary tree with $n$ leaves (and therefore $\log_k(n)$
depth) structured as follows. We start at the root, then recursively define the
rest of the tree. Every node in the tree represents a single optimizer state.
The root represents state $s_0$. To recursively define the remaining nodes: each
non-leaf node $s_i$ at depth $d$ has $k$ equally spaced (in terms of state
number) children starting---from left to right---at state $s_i$ and ending at
$s_{i + n/k^{d+1}}$. This means that the leaves correspond---from left to
right---to the states $s_0, s_1, \ldots, s_{n-1}$.

We reduce the problem of iterating over the states in reverse to the problem of
reverse in-order traversing this tree and yielding \textit{just} the
leaves---this is exactly the states in reverse order. A reverse in-order
traversal for this $k$-ary tree requires repeatedly: recursively traversing
child nodes from largest to smallest, then visiting the parent node. We design
the specifics of this traversal to maximize space and compute efficiency. To
access the children of a parent node at traversal time, we replay model training
from the smallest child state (which is stored in the parent state) to the
largest child state and store all the children. We perform this operation
recursively each time we traverse a node. After traversing the node's left side
(i.e., after ascending from this node), we delete all its child states.

Reverse in-order traversing this tree requires storing at most $k\log_k(n)$
optimizer states at a time, and in aggregate requires retraining the model
$\log_k(n)$ times. The argument for each is straightforward. Storage: the
traversal requires storing at most $k$ states for each level that it descends
(we store $k$ states whenever we first traverse to a parent node) and we remove
$k$ states for each level that the traversal ascends (we remove $k$ states after
we are done with the left traversal of a parent). Compute: we replay training to
reinstantiate the children of every parent node a single time. The $k^d$ parent
nodes at level $d$ each require replaying $\mathcal{O}(\nicefrac{n}{k^d})$
states to reinstantiate children. Therefore, in a traversal, each level requires
$\mathcal{O}(n)$ ($k^d \cdot \nicefrac{n}{k^d}$) optimizer steps. There are
$\log_k(n)$ levels with parent nodes, which means a total of $\mathcal{O}(n
\log_k(n))$ optimizer steps, or a multiplicative factor of
$\mathcal{O}(\log_k(n))$ steps compared to model training.

\begin{algorithm}
  \SetKwFor{For}{\textcolor{forcolor}{for}}{\textcolor{forcolor}{do}}{}
  \SetKw{Return}{\textcolor{forcolor}{Return}}{}
  \SetKwFunction{traversereverse}{\textcolor{fncolor}{reverse\_inorder\_traversal}}
  \DontPrintSemicolon $T \gets $ Lazy $k$-ary tree for $\mathcal{A}(z)$ \qquad \tcp{Make
  lazy $k$-ary tree of Appendix~\ref{app:replay}} \;

  \hspace{0pt} \tcp{Variables; shorthand for $\smash{\frac{\partial f(z)}{\partial z}}$ and $\smash{\frac{\partial f(z)}{\partial s_T}}$}
  $\bar{z} \gets 0$ \;
  $\bar{s}_T \gets \smash{\pfrac{g(s_T)}{s_T}}$ \qquad \tcp{One reverse-mode AD call} %
  \;
  \hspace{0pt} \tcp{Reverse-mode differentiate step-by-step; \underline{traverse $T$ instead of stored states}}
  \For{$s_i \gets s_{T-1}$ \color{forcolor}{\textbf{to}} \color{black}$s_0 \in \traversereverse{T}$}{ 
  \hspace{0pt} \tcp{One reverse-mode AD call. Left: $\nabla_{s_i} f$. Right: contribution to $\nabla_{z} f$ at $i$.}
      $\bar{s}_i \gets \bar{s}_{i+1} \cdot \pfrac{h_{i}(s_i, z)}{s_i}, \qquad \bar{z}_i \gets \bar{s}_{i+1} \cdot \pfrac{h_{i}(s_i, z)}{z}$\;
      \;
      $\bar{z} \leftarrow \bar{z} + \bar{z}_i$ \qquad \tcp{Accumulate metagradient}
  }
  \;
  \Return $\bar{z}$ \;
  \caption{\textsc{Replay}. metagradients in
  $\mathcal{O}(k\log_k(T))$ space.}
  \label{alg:replay}
\end{algorithm}

%% file: sections/app/clip_details.tex
This appendix contains pseudocode for the main algorithm used to do dataset
selection for DataComp. It also contains additional implementation details on
how metagradients were applied to CLIP, and how they were specifically applied
to the DataComp setting.

\subsection{Dataset Selection Using MGD}

When implementing Algorithm \ref{alg:depsing}, there are several differences from the pseudocode below:
firstly, rather than selecting $\mathbf{m}$ fully randomly every step, we
randomly select a shard comprising fraction $p$ of the data and take steps on
all datapoints in the shard (see Section~\ref{sec:scaling_clip}).  To mitigate
overfitting, we also bake a ``minibatch fraction'' $q$ into our model output
function $\phi$.  For example, if $\phi$ calculates model loss on the ImageNet
train set, each time $\phi$ is called, we randomly sample fraction $q$ of the
ImageNet train set to evaluate on.

\paragraph{Adapting the CLIP loss function to our surrogate learning algorithm.}
Here, we explain how dataweights are incorporated into the CLIP loss
function---the formulation given in Section \ref{sec:datacomp} is actually
slightly simplified and incorrect, as it does not account for cross terms in
the CLIP contrastive loss.  As a refresher, we first state the ``vanilla'' CLIP
loss function, $\ell$, as it is defined in \cite{radford2021learning}.  Letting
$b$ be the batch size and $d$ be the embedding dimension, and $\mathbf{x}$ be
the
training batch at timestep $k$.
Recall that the CLIP model internally has two ``submodules'': and image
embedder, and a text embedder.  We then use these to obtain image embeddings
$E_I \in \mathbb{R}^{b \times d}$ and text embeddings $E_T \in
\mathbb{R}^{b \times d}$ from $\mathbf{x}$. We then compute the image-wise
scores, or logits, for this batch as
$S = E_I E_T^\top$ \footnote{The CLIP model scales these logits by a
	temperature parameter $\tau$ before applying the softmax. While we omit
	$\tau$ in our definitions, it can be easily incorporated. All our
experiments use temperature scaling.}.
Then, we can define the CLIP loss (as a function of the logits) as
\begin{equation*}
	L(S) = \frac{1}{2} (L_I(S) + L_T(S)),
\end{equation*}
where $L_I$ and $L_T$ are row-wise and column-wise cross-entropy losses, respectively:
\begin{equation*}
	L_I(S) = \sum_{i=1}^b 
    \log \left(
      \frac{\exp(S_{i,i})}
           {\sum_{j=1}^b \exp(S_{i,j})}
    \right), \quad
	L_T(S) = \sum_{i=1}^b 
    \log \left(
      \frac{\exp(S_{i,i})}
           {\sum_{j=1}^b \exp(S_{j,i})}
    \right).
\end{equation*}

We now wish to relax $L$ into a new function $L'$ that supports an additional
input $\mathbf{z} \in \mathbb{R}^n$, where $\frac{\partial L'}{\partial
\mathbf{z}}$ resembles the metagradients with respect to dataweights.  In order
to do this, we imagine expanding passing the \emph{entire dataset} $D$ into our
embedder to obtain $E_I'$ and $E_T'$, and take our new logits $S' = E_I'
E_T'^\top \in \mathbb{R}^{n \times n}$.

There are some additional key conditions our relaxation $L'$ should satisfy. 
Particularly:  when $\mathbf{z} = \mathbf{0}_n$, we should recover the normal
CLIP loss $L$, and when $\mathbf{z}$ is all 0's except for a single entry $i$,
$L'$ should act as if $i$ had been appended to the original batch $\mathbf{x}$.
In addition, $L'$ should always have meaningful partials with
respect to $\mathbf{z}$, even when some values in $\mathbf{z}$ are 0.  

Letting $\mathbf{1}_{i = j}$ and $\mathbf{1}_{i \neq j}$ be indicator variables
and letting $\mathbf{1}_{k} \in \{0, 1\}^n$ be the indicator vector for the
$k$-th batch, we find that the definition
\[
	L'(S', \mathbf{z}) = L'_I(S', \mathbf{z}) + L'_T(S', \mathbf{z}),
\]
where
\[
	L'_I(S', \mathbf{z}) = \sum_{i=1}^n
	(z_i + (\mathbf{1}_{k})_i) \log \left(
      \frac{\exp(S'_{i,i})}
	  {\sum_{j=1}^n \exp(S'_{i,j}) \left(\mathbf{1}_{i = j} + \mathbf{1}_{i \neq j} (z_j + (\mathbf{1}_k)_j)\right)}
    \right)
\]
and
\[
	L'_T(S', \mathbf{z}) = \sum_{i=1}^b 
	(z_i + (\mathbf{1}_{k})_i) \log \left(
      \frac{\exp(S'_{i,i})}
	  {\sum_{j=1}^n \exp(S'_{j,i}) \left(\mathbf{1}_{i = j} + \mathbf{1}_{i \neq j} (z_j + (\mathbf{1}_k)_j)\right)}
    \right)
\]
satisfy these conditions.  

Finally, we let define the loss for the entire batch $\ell'$ as a function of $\mathbf{z}$ and model parameters $\theta$ which outputs the loss calculated according to $L'$ above.
To summarize, letting $\mathbf{x}^{(t)}$ denote the $t$-th training batch, the loss function $\ell_t$ at step $t$ of our surrogate learning algorithm $\mathcal{A}'$ for CLIP training is:
\begin{equation*}
	\label{eq:surrogate_objective_new}
    \ell'_t(\theta) := 
    \begin{cases}
		\ell(\mathbf{x}^{(t)}; \theta) & \text{if } t \neq k \\
		\ell'(\mathbf{z}; \theta) & \text{if } t = k.
    \end{cases}
\end{equation*}
We find that this empirically works well for obtaining meaningful
metagradients with respect to dataweights in the CLIP setting, and yields to strong dataset selection results.

\subsection{Scaling MGD for CLIP and DataComp}
\label{sec:scaling_clip}

MGD is highly scalable, allowing it to be applied to large-scale settings like
training CLIP models.  In particular, computing metagradients is only up to a
constant factor more expensive than training a model normally. 
Here, we outline challenges we faced in scaling MGD in this setting, and how they were resolved.
Specifically, we will explain how we efficiently calculated metagradients for
CLIP models and efficiently tracked/shuffled our dataset selection from
step-to-step despite its large storage footprint.

\paragraph{Computing metagradients.}
Due to the large batch size used in the CLIP contrastive loss, we implement
manual gradient checkpointing to make the operations computationally feasible
on our hardware.  The most memory-intensive operation are model forward passes
(and its gradients): obtaining the image and label embeddings given raw pixel
data and tokens.  So, we manually make gradient checkpoints before this
operation, allowing us to run the embedder in minibatches to avoid memory
issues.  This setup also naturally lends itself to parallelization across
multiple GPU's, which we make use of to further speed up our computations.

\paragraph{Loading, writing, and storing data.}
Due to the data-intensive nature of training large models like CLIP and our
need to frequently produce new datasets at each optimization step, we found
that using the webdataset \cite{webdataset2024webdataset} format given by
DataComp was restrictively slow.  To circumvent this, we rewrote all data
following the format of FFCV \cite{leclerc2022ffcv}, allowing us to load and
write data much faster.  Specifically, we divided the entire candidate pool
into 8 base shards.  Once we trained a model, we choose one of the 8 shards,
compute metagradients corresponding to all datapoints in the shard, take a
gradient step on them, and rewrite the shard.  This roughly corresponds to $p =
\frac{1}{8}$ in Algorithm \ref{alg:depsing}, which we empirically worked well
for optimizing.  In following steps, we always choose one of the 8
\emph{original} shards to calculate metagradients for---this ensures that
points removed from the dataset in some optimization step can return if they
have a negative metagradient.

We also observed that always stepping on the sign causes the sizes of the
shards to grow over time: stepping based on the sign of the metagradient does
not decrease the weight on a positive-weight datapoint if its dataweight  is
already 0, so our steps are biased towards increasing the size of the shards. 
To combat this blowup, after some number of optimization steps, we choose a
fixed shard size and enforce that subsequent steps must not change the size of
the shards---the step size thereafter is controlled by hyperparameter $q$
representing the fraction of datapoints in a shard which are incremented.  We
experimented both with randomly sampling which points are added or removed, and
stepping on the datapoints with the top $q$ and bottom $q$ metagradients; the 
latter seems to give empirically better performance.

To maintain randomness during shuffling, we implement an 8-way dataloader which
would shuffle all 8 shards individually.  Then, to sample a batch of $b$
datapoints, we would sample $b / 8$ datapoints from each shard and concatenate
them to fill our batch.  This works better than simply sampling our entire
batch from a single shard, as (especially in later optimization steps) shards
may contain a high number of duplicate datapoints, which causes CLIP's
contrastive loss function to misbehave if they appear in the same batch.

To minimize disk space used, old shards can be deleted once they become
``stale''.  Specifically, if shard $s$ is rewritten into shard $s'$, all future
optimization steps will never read $s$ again, and $s$ can safely be
deleted.  Thus, when running MGD for a large number of steps and potentially
rewriting each shard multiple times, the total disk space used by our algorithm
is constant in the number of steps we take: it stores the 8 most recently written 
shards on disk at any given time, and any other shards are deleted to save space.

\subsection{Details Pertaining to the DataComp Benchmark}

\paragraph{Setting.}
We provide a brief summary of the DataComp competition here, and we refer
readers to the original paper \cite{gadre2024datacomp}.  DataComp is a
framework to compare different training dataset selection techniques. 
Participants submit a training dataset (which, for our purposes, is a subset of
a larger dataset), upon which a CLIP model is trained from scratch with a fixed
learning algorithm, model architecture, and number of training steps.  We focus
on DataComp-small, which has a candidate pool of 12.8 million samples.  The
number of training steps in this case is also fixed at 12.8 million samples. 

We try to match the optimization hyperparameters enforced by DataComp as
closely as possible.  As a refresher, our ADAM\cite{kingma2015adam} update step
can be written as
\begin{equation}
	\label{eq:adam_update}
	\theta_{t+1} = -\alpha_t \cdot \left(m_t / \left(\sqrt{v_t +
	\epsilon_\mathrm{root}} + \epsilon\right) + \lambda \theta_t\right)
\end{equation}
where $m_t$ and $v_t$ are running estimates of the first and second moments of
the gradients, respectively, $\lambda$ represents weight decay, $\alpha$
represents the learning rate, and $\epsilon$ and $\epsilon_\mathrm{root}$ are
hyperparameters to avoid blowup.  Our training hyperparameters can be found in
Table \ref{tab:clip_hparams} and are identical to those mandated by
DataComp-small, aside from a positive $\epsilon_\mathrm{root}$ added for
numerical stability. The values of $\epsilon_{\mathrm{root}}$ and $k$ (the step
at which metagradients are calculated) were chosen to empirically maximize
metasmoothness.

\begin{table}[h]
	\caption{Hyperparameters for the CLIP DataComp experiments.}
	\centering
	\input{tables/clip_hparams_table.tex}
	\label{tab:clip_hparams}
\end{table}

Our experiments are also run on an incomplete subset of the entire DataComp
candidate pool.    DataComp did not store the raw image and text files when
assembling their dataset; they only stored a list of URL's to download data
from.  Due to the nature of the internet, for various reasons, some of these
URL's no longer point to the same data (or no longer point to any data at all).
Thus, after ignoring these broken links, our candidate pool is only around
$80\%$ of the size of the original DataComp candidate pool when it was
collected in 2023.  All our results are obtained by running our methods on this
subset of the DataComp pool.

\paragraph{Evaluation tasks.}
In order to ensure that our method is truly improving trained models'
performances on the \emph{entire target distribution} and not overfitting to
the target set, for each of the 38 evaluation tasks used by DataComp, we
attempted to separately create a disjoint target and validation set (DataComp only
creates test sets for each task).  Thus, metagradients were computed on the
target sets and model performance was evaluated on the validation set, before
submitting with the official DataComp script and evaluating on the test sets. 
This ensures that our method's generalization ability is being evaluated, and
we are not overfitting to our target set.

For various reasons, creating target splits was not possible for all 38 tasks;
we summarize our setup in Table \ref{tab:datacomp_tasks}.

{
\renewcommand{\thefootnote}{\fnsymbol{footnote}}
\begin{table}[h]
\caption{All DataComp evaluation tasks.  The ``Target set'' column refers to whether metagradients were taken on the target set corresponding to this dataset.}
\centering
{
	\scriptsize
	\input{tables/clip_tasks_table.tex}
}
\label{tab:datacomp_tasks}
\end{table}
\footnotetext[1]{No train or val set exists for this dataset, so we were unable to create disjoint target and val sets.}
\footnotetext[2]{We were unable to use this dataset due to technical difficulties.}
\footnotetext[3]{Both the train and val sets were used by DataComp to make their test set, so we were unable to create disjoint target and val sets.}
\footnotetext[4]{Retrieval tasks were not used for metagradients.}
}

%% file: tables/clip_hparams_table.tex
\begin{tabular}{lc}
\toprule
\textbf{Hyperparameter} & \textbf{Value} \\
\midrule
DataComp Scale & small \\
Model & ViT-B/32 \\
Train compute (MACs) & $9.5 \times 10^{16}$ \\
Pool size & 12.8M \\
\# samples seen & 12.8M \\
Batch size & 4096 \\
Training batches & 3125 \\
$k$ & 2800 \\
Learning rate & $5 \times 10^{-4}$ \\
AdamW $\beta_1$ & 0.9 \\
AdamW $\beta_2$ & 0.98 \\
AdamW $\epsilon_{\textrm{root}}$ & $1 \times 10^{-17}$ \\
Warmup & 500 \\
\bottomrule
\end{tabular}

%% file: tables/clip_tasks_table.tex
\begin{tabular}{llllllc}
\toprule
\textbf{Dataset}  &  \textbf{Task}  &  \textbf{Test size}  &  \textbf{Train size}  &  \textbf{Val size}  &  \textbf{Main metric}  &  \textbf{Target set} \\
\midrule
Caltech-101 \cite{fei2004learning}  &  Object recognition  & 6085 &  2754  &  306  &  mean per class  &  \checkmark \\
CIFAR-10 \cite{krizhevsky2009learning}  &  Visual recognition  & 10000 &  45000  &  5000  &  accuracy  &  \checkmark \\
CIFAR-100 \cite{krizhevsky2009learning}  &  Visual recognition  & 10000 &  45000  &  5000  &  accuracy  &  \checkmark \\
CLEVR Counts \cite{johnson2017clevr,zhai2019large}  &  Counting  & 15000 &  65000  &  5000  &  accuracy  &  \checkmark \\
CLEVR Distance \cite{johnson2017clevr,zhai2019large}  &  Distance prediction  & 15000 &  65000  &  5000  &  accuracy  &  \checkmark \\
Country211 \cite{radford2021learning,thomee2016yfcc100m}  &  Geolocation  & 21100 &  37980  &  4220  &  accuracy  &  \checkmark \\
DTD \cite{cimpoi2014describing}  &  Texture classification  & 1880 &  3384  &  376  &  accuracy  &  \checkmark \\
EuroSAT \cite{helber2019eurosat,zhai2019large}  &  Satellite imagery recognition  & 5400 &  19440  &  2160  &  accuracy  &  \checkmark \\
FGVC Aircraft \cite{maji2013fine}  &  Aircraft recognition  & 3333 &  6001  &  666  &  mean per class  &  \checkmark \\
Food-101 \cite{bossard2014food}  &  Food recognition  & 25250 &  70750  &  5000  &  accuracy  &  \checkmark \\
GTSRB \cite{stallkamp2011german}  &  Traffic sign recognition  & 12630 &  35289  &  3920  &  accuracy  &  \checkmark \\
ImageNet 1k \cite{deng2009imagenet}  &  Visual recognition  & 50000 &  1276167  &  5000  &  accuracy  &  \checkmark \\
ImageNet Sketch \cite{wang2019learning}  &  Visual recognition  & 50889 &  N/A  &  N/A  &  accuracy  &   \footnotemark[1] \\
ImageNet V2 \cite{recht2019imagenet}  &  Visual recognition  & 10000 &  N/A  &  N/A  &  accuracy  &   \footnotemark[1] \\
ImageNet-A \cite{hendrycks2019natural}  &  Visual recognition  & 7500 &  N/A  &  N/A  &  accuracy  &   \footnotemark[1] \\
ImageNet-O \cite{hendrycks2019natural}  &  Visual recognition  & 2000 &  N/A  &  N/A  &  accuracy  &   \footnotemark[1] \\
ImageNet-R \cite{hendrycks2020faces}  &  Visual recognition  & 30000 &  N/A  &  N/A  &  accuracy  &   \footnotemark[1] \\
KITTI distance \cite{geiger2012ready,zhai2019large}  &  Distance prediction  & 711 &  N/A  &  N/A  &  accuracy  &   \footnotemark[2] \\
MNIST \cite{lecun1998mnist}  &  Digit recognition  & 10000 &  55000  &  5000  &  accuracy  &  \checkmark \\
ObjectNet \cite{barbu2019objectnet}  &  Visual recognition  & 18574 &  N/A  &  N/A  &  accuracy  &  \footnotemark[1] \\
Oxford Flowers-102 \cite{nilsback2008automated}  &  Flower recognition  & 6149 &  1836  &  204  &  mean per class  &  \checkmark \\
Oxford-IIIT Pet \cite{parkhi2012cats,zhai2019large}  &  Pet classification  & 3669 &  3312  &  368  &  mean per class  &  \checkmark \\
Pascal VOC 2007 \cite{everingham2010pascal}  &  Object recognition  & 14976 &  14096  &  1566  &  accuracy  &  \checkmark \\
PatchCamelyon \cite{veeling2018rotation,zhai2019large}  &  Metastatic tissue cls.  & 32768 &  289912  &  5000  &  accuracy  &  \checkmark \\
Rendered SST2 \cite{zhai2019large}  &  Sentiment classification  & 1821 &  7013  &  779  &  accuracy  &  \checkmark \\
RESISC45 \cite{cheng2017remote,zhai2019large}  &  Satellite imagery recognition  & 6300 &  22680  &  2520  &  accuracy  &  \checkmark \\
Stanford Cars \cite{krause20133d}  &  Vehicle recognition  & 8041 &  7329  &  814  &  accuracy  &  \checkmark \\
STL-10 \cite{coates2011analysis}  &  Visual recognition  & 8000 &  4500  &  500  &  accuracy  &  \checkmark \\
SUN-397 \cite{xiao2010sun}  &  Scene recognition  & 108753 &  N/A  &  N/A  &  accuracy  &  \footnotemark[3] \\
SVHN \cite{netzer2011reading,zhai2019large}  &  Digit recognition  & 26032 &  68257  &  5000  &  accuracy  &  \checkmark \\
iWildCam \cite{beery2021iwildcam,koh2020wilds}  &  Animal recognition  & 42791 &  147084  &  5000  &  macro F1 score  &  \checkmark \\
Camelyon17 \cite{bandi2018from,koh2020wilds}  &  Metastatic tissue cls.  & 85054 &  365900  &  5000  &  accuracy  &  \checkmark \\
FMoW \cite{christie2018functional,koh2020wilds}  &  Satellite imagery recognition  & 22108 &  103261  &  5000  &  worst-region acc.  &  \checkmark \\
Dollar Street \cite{rojas2022dollar}  &  Object recognition  & 3503 &  13842  &  1537  &  worst-income top-5 acc.  &  \checkmark \\
GeoDE \cite{ramaswamy2024geode}  &  Object recognition  & 12438 &  44488  &  4943  &  worst-region acc.  &  \checkmark \\
Flickr30k \cite{young2014from}  &  Image and text retrieval  & 31014 &  N/A  &  N/A  &  R@1  &  \footnotemark[4] \\
MSCOCO \cite{lin2014microsoft}  &  Image and text retrieval  & 5000 &  N/A  &  N/A  &  R@1  &  \footnotemark[4] \\
WinoGAViL \cite{bitton2022winogavil}  &  Commonsense association  & 3563 &  N/A  &  N/A  &  Jaccard score  &  \footnotemark[4] \\
\bottomrule
\end{tabular}

%% file: sections/app/ift_details.tex
In this section, we describe the details of the IFT setting of \citet{xia2024less},
as well as the details of our method.

\paragraph{Setting.}
The setting contains a fixed data pool: instruction fine-tuning data from a data
pool consisting of four combined IFT datasets (cf. Table~\ref{tab:ift_training_dataset}
and \citet{xia2024less} for more information). The goal is to select the data
that yields the best possible task performance for a LoRA fine-tuning run.
We adapt a LoRA to a Gemma-2B model (the pretraining-only Gemma-2B model)
using the LoRA configuration from \citet{xia2024less}.

\paragraph{Data splits.}
See Table~\ref{tab:datasets} for a description of the available data for each
task, along with the task setup details. \citet{xia2024less} constructed these
extra samples by drawing from the ICL samples given in the tasks originally.
Note that we drop TydiQA from the original work of \citet{xia2024less} as there
are not enough samples to select with (there is only one from each category, for
a total of 7). 

\paragraph{Method.}
We execute Algorithm~\ref{alg:depsing} with $k$ as 150 steps from the end of training and
the Bernoulli parameter $p$ controlling the step size as 0.2. At each step, we
choose a ``minibatch'' with a size equal to half the target set and a quarter of
the target set for BBH and MMLU, respectively (that is, we only select to
optimize performance on a fraction of the target set at a time). We model
select over iterates and hyperparameters by (a) choosing the top three steps in
terms of validation loss for each run (b) selecting the best one in terms of
full train set accuracy (including the part that we trained on). We perform this
procedure---akin to Pareto optimization~\citep{jin2008pareto}---because the
validation set is so small (as the overall set of samples is very small) that it
is difficult to select models without overfitting otherwise.

We compare with two baselines: training on the full dataset (i.e., training on
the entirety of all the data for a single epoch), and LESS (we use the data
selected according to ``LESS-T''~\citep{xia2024less}, following the
recommendation of 4 epochs).

For model training, we train with ADAM ($\beta_1=0.95, \beta_2=0.975$,
decoupled weight decay as $10^{-5}$) and a one-cycle linear schedule starting
at $10^{-6}$ of the maximum learning rate, reaching the peak over 25\% of
training, then ending at 0.1 of the maximum learning rate. We insert a positive
$\epsilon_\textrm{root}$ into the inverse square root term in the ADAM update
to prevent metagradient (and to a lesser extent update) blowup (see Eq.\
\ref{eq:adam_update}). The model training is the same across selected data,
except that we use $\epsilon_\textrm{root}=10^{-7}$ for MGD-selected data and
$\epsilon_\textrm{root}=10^{-9}$ for the other runs (we select the optimal
parameter for each class of method). We additionally hyperparameter select for
the best learning rate across each baseline by minimizing validation set loss;
LESS performs best with a smaller learning rate ($0.00024$ for BBH and
$0.00012$ for MMLU) than training on the full dataset or with MGD ($0.0006$ for
both). We normalize the loss of each training sample by taking the mean across
predicted tokens during training, and do not divide by the batch size
(important for scaling the $\epsilon_\mathrm{root}$ term, but otherwise ADAM is
invariant to the scale). 

\paragraph{Selecting smooth model training for MGD.} For MGD runs, we jointly
select learning rate and $\epsilon_{\mathrm{root}}$ using the smoothness metric
of Section~\ref{sec:localpred}. We find that the choice of $\epsilon_\mathrm{root}$ term
is important (just as the choice of $\epsilon$ is important in standard ADAM training); choosing a much larger term results in non-smooth training. We also find that
metagradients are sensitive to learning rate schedule; choosing a much larger or
smaller maximum learning rate results in non-smooth training.

\begin{table}[h]
    \caption{Overview of datasets used in IFT dataset selection (from \citet{xia2024less}).}
    \label{tab:datasets}
    \centering
    \begin{tabular}{lccccccc}
        \toprule
        \textbf{Dataset} & \textbf{\# Shot} & \textbf{\# Tasks} & $n_\mathrm{target}$ & $n_\mathrm{val}$  & $n_\mathrm{test}$ & \textbf{Answer Type} & \textbf{Type of Task} \\
        \midrule
        MMLU & 5 & 57 & 57 & 228 & 18,721 & Letter options & Knowledge/Recall \\
        BBH & 3 & 23 & 23 & 46 & 920 & COT and answer & Reasoning \\
        \bottomrule
    \end{tabular}
\end{table}

\begin{table}[h]
    \centering
    \caption{Details of IFT training datasets.}
    \label{tab:ift_training_dataset}
    \begin{tabular}{l l l c c}
        \toprule
        Dataset & \# Instance & Sourced from & Prompt Len. & Completion Len. \\
        \midrule
        FLAN V2 & 100,000 & NLP datasets and human-written instructions & 355.7 & 31.2 \\
        CoT & 100,000 & NLP datasets and human-written CoTs & 266 & 53.2 \\
        Dolly & 15,011 & Human-written from scratch & 118.1 & 91.3 \\
        Open Assistant 1 & 55,668 & Human-written from scratch & 34.8 & 212.5 \\
        \bottomrule
    \end{tabular}
\end{table}

\paragraph{IFT results} \quad %

\begin{figure}[hb]
	\centering
	\caption{MGD dataset selection improves the validation loss over metagradient steps, demonstrating our method's efficacy.  However, the gap between loss on samples MGD directly optimizes on and the validation samples widens over the number of iterates, and there is overfitting depending on the number of steps taken. }
	\includegraphics[width=0.45\textwidth]{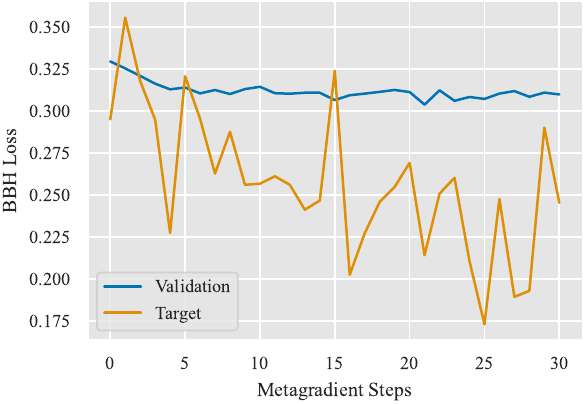}
	\quad
	\includegraphics[width=0.45\textwidth]{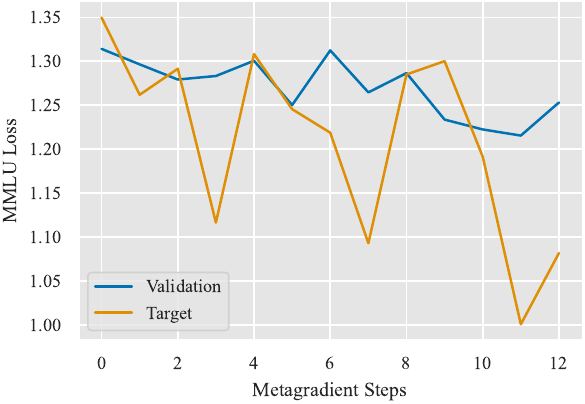}
	\label{fig:bbh_mmlu_over_time}
\end{figure}

%% file: tables/cifar_hparams.tex
\begin{tabular}{lc}
\toprule
\textbf{Hyperparameter} & \textbf{Value} \\
\midrule
Learning rate & $0.5$ \\
$\beta_1$ & $0.85$ \\
Weight decay & $10^{-5}$ \\
Exclude BatchNorm & True \\
Optimizer & SGD \\
Batch size & $250$ \\
Epochs & $18$ \\
Starting learning rate fraction & $0.5$ \\
Relative min. learning rate & $10000$ \\
Scheduler max. iterations & $50000$ \\
Nesterov momentum & True \\
BatchNorm $\epsilon$ & $10^{-5}$ \\
BatchNorm momentum & $0.5$ \\
Final bias & True \\
Width multiplier & $2.0$ \\
Final scale & $0.125$ \\
Initial scale & $2.0$ \\
Batchnorm location & Before activation \\
Activation function & GELU \\
Pooling type & Average \\
Test-time augmentation & True \\
\bottomrule
\end{tabular}

%% file: pgffigs/lr_sched_plot/sched_plot.tex
\pgfplotsset{colormap/Set2}
\pgfplotsset{scaled y ticks=false}
\pgfplotsset{grid style={dashed,gray}}

\definecolor{mycolor0}{HTML}{66C1A5}
\definecolor{mycolor1}{HTML}{6EBAAD}
\definecolor{mycolor2}{HTML}{75B4B4}
\definecolor{mycolor3}{HTML}{7DADBC}
\definecolor{mycolor4}{HTML}{84A7C3}
\definecolor{mycolor5}{HTML}{8CA0CB}

\begin{tikzpicture}
    \begin{axis}[
      xlabel={Training step},
	  ylabel={Learning rate},
      height={7cm},
      width={10cm},
      colormap name={Set2},
	  legend style={
			title=MGD step, %
            at={(1.02,1)},      %
            anchor=north west,  %
	},
      grid=both
    ]
	\addplot[
	  color=mycolor0,
	  mark size=0.0,
	  very thick,
      error bars/.cd,
      y dir=both,
      y explicit,
    ] table[
      x=x,
      y=1,
      col sep=comma   %
	]{pgffigs/lr_sched_plot/scheds.csv};
	\addlegendentry{1}

	\addplot[
	  color=mycolor1,
	  mark size=0.0,
	  very thick,
      error bars/.cd,
      y dir=both,
      y explicit,
    ] table[
      x=x,
      y=11,
      col sep=comma   %
	]{pgffigs/lr_sched_plot/scheds.csv};
	\addlegendentry{11}

	\addplot[
	  color=mycolor2,
	  mark size=0.0,
	  very thick,
      error bars/.cd,
      y dir=both,
      y explicit,
    ] table[
      x=x,
      y=21,
      col sep=comma   %
	]{pgffigs/lr_sched_plot/scheds.csv};
	\addlegendentry{21}

	\addplot[
	  color=mycolor3,
	  mark size=0.0,
	  very thick,
      error bars/.cd,
      y dir=both,
      y explicit,
    ] table[
      x=x,
      y=31,
      col sep=comma   %
	]{pgffigs/lr_sched_plot/scheds.csv};
	\addlegendentry{31}

	\addplot[
	  color=mycolor4,
	  mark size=0.0,
	  very thick,
      error bars/.cd,
      y dir=both,
      y explicit,
    ] table[
      x=x,
      y=41,
      col sep=comma   %
	]{pgffigs/lr_sched_plot/scheds.csv};
	\addlegendentry{41}

	\addplot[
	  color=mycolor5,
	  mark size=0.0,
	  very thick,
      error bars/.cd,
      y dir=both,
      y explicit,
    ] table[
      x=x,
      y=51,
      col sep=comma   %
	]{pgffigs/lr_sched_plot/scheds.csv};
	\addlegendentry{51}
    \end{axis}
\end{tikzpicture}